\documentclass{classes/myifcolog}
\usepackage[utf8]{inputenc}

\usepackage{classes/llncsdoc}
\usepackage{proof}
\usepackage{latexsym}
\usepackage{amssymb,amsmath,stmaryrd}
\usepackage{mathtools}
\usepackage{rotating}

\usepackage{epsf}
\usepackage[all]{xy}
\usepackage{lscape}
\usepackage{verbatim}
\usepackage{enumitem}
\usepackage{times}
\usepackage{color}

\usepackage{lingmacros}

\newcommand{\cut}[1]{}
\newcommand{\W}[1]{\textrm{#1}}
\newcommand{\bs}{\backslash}
\newcommand{\Ra}{\rightarrow}
\newcommand{\fdia}{\diamondsuit}
\newcommand{\rdia}{\textcolor{red}{\fdia}}
\newcommand{\gbox}{\Box}
\newcommand{\rbox}{\textcolor{red}{\gbox}}
\newcommand{\hm}[1]{\lceil #1 \rceil}
\newcommand{\ov}{\overrightarrow}

\newcommand{\arrw}[2]{#1\longrightarrow #2}

\newcommand{\tensor}{\otimes}
\newcommand{\s}{\slash}
\newcommand{\arr}[3]{#1:#2\longrightarrow #3}

\newcommand{\pijl}{\rightarrow}

\newcommand{\xrighta}{\alpha_{\fdia}^{}}
\newcommand{\xrightc}{\sigma_{\fdia}^{}}

\newcommand{\modalNL}{\textbf{NL}$_{\fdia}$}
\newcommand{\catvect}{\textbf{FVect}}

\newcommand{\gap}{\textvisiblespace}
\usepackage{tikz}
\usepackage{pgfplots}
\usetikzlibrary{trees}
\usetikzlibrary{topaths}
\usetikzlibrary{decorations.pathmorphing}
\usetikzlibrary{decorations.markings}
\usetikzlibrary{matrix,backgrounds,folding}
\usetikzlibrary{chains,scopes,positioning,fit,shapes}
\usetikzlibrary{arrows,shadows}
\usetikzlibrary{arrows.meta}
\usetikzlibrary{calc} 
\usetikzlibrary{chains}
\usetikzlibrary{shapes,shapes.geometric,shapes.misc}

\pgfdeclarelayer{edgelayer}
\pgfdeclarelayer{nodelayer}
\pgfsetlayers{background,edgelayer,nodelayer,main}

\tikzstyle{dot}=[circle,fill=black,draw=black]
\tikzstyle{every picture}=[baseline=(current bounding box).east,scale=0.5,node distance=5mm]
\tikzstyle{none}=[inner sep=0pt]
\tikzstyle{every loop}=[]

\tikzstyle{(null)}=[]
\tikzstyle{plain}=[]

\tikzstyle{tensornode}=[circle,draw=black,inner sep=.5mm]
\tikzstyle{parnode}=[circle,draw=black,fill=black!20,inner sep=.5mm]
\tikzstyle{graphArrow}=[thick,-{Triangle[length=1mm,width=1.5mm]}]
\tikzstyle{graphNoArrow}=[thick]

\titlethanks{}
\title{A Frobenius Algebraic Analysis\\ for Parasitic Gaps}
\titlerunning{A Frobenius Algebraic Analysis for Parasitic Gaps}
\addauthor[m.j.moortgat@uu.nl]{Michael Moortgat}{Utrecht University\thanks{The research of the alphabetically first and third author is supported by NWO grant 360-89-070 ``A composition calculus for vector-based semantic modelling with a localization for Dutch'' .}}
\addauthor[m.sadrzadeh@ucl.ac.uk]{Mehrnoosh Sadrzadeh}{University College London\thanks{The alphabetically second author acknowledges the support of Royal Society International Exchange Award IE161631.}}
\addauthor[g.j.wijnholds@uu.nl]{Gijs Wijnholds}{Utrecht University}
\authorrunning{Moortgat, M., Sadrzadeh, M., and Wijnholds, G.}
\date{\today}

\begin{document}
\maketitle

\begin{abstract}
The interpretation of parasitic gaps is an ostensible case of non-linearity in natural language composition.
Existing categorial analyses, both in the typelogical and in the combinatory traditions, rely on explicit
forms of syntactic copying. We identify two types of parasitic gapping where the duplication of
semantic content can be confined to the lexicon. Parasitic gaps in \emph{adjuncts} are analysed as
forms of generalized coordination with a polymorphic type schema for the head of the adjunct phrase.
For parasitic gaps affecting \emph{arguments} of the same predicate, the polymorphism is associated
with the lexical item that introduces the primary gap. Our analysis is formulated in terms of
Lambek calculus extended with structural control modalities. A compositional translation
relates syntactic types and derivations to the interpreting compact closed category of finite
dimensional vector spaces and linear maps with Frobenius algebras over it.
When interpreted over the necessary semantic spaces, the Frobenius algebras
provide the tools to model the proposed instances of lexical polymorphism.
\end{abstract}

\section{Introduction}

Natural languages present many situations where an overt syntactic element provides the
semantic content for one or more occurrences of elements that are not physically realized,
or that have no meaning of their own.
Illustrative cases can be found at the sentence and at the discourse level, 
e.g.~long-distance dependencies in `movement' constructions, ellipsis phenomena, anaphora.
Parasitic gaps are a challenging case in point.

To provide the reader with the necessary linguistic background, the examples in (\ref{data})
illustrate some relevant patterns\footnote{For a more thorough discussion of the phenomena, and proposed
analyses in a variety of grammatical frameworks, see\cite{parasitic2001}.}. 
The symbol \textvisiblespace\ marks the position of
the virtual elements that depend on a physically realized phrase elsewhere in their  context for their interpretation;
in the generative grammar literature, these virtual elements are referred to as ``gaps''.

%As the name suggests, a \emph{parasitic} gap is felicitous only in the presence of a primary gap.
%The examples in (\ref{data}) illustrate some relevant patterns.

\begin{equation}\label{data}
\mbox{\begin{tabular}{r@{\quad}r@{}l@{\qquad}l}
$a$ & & papers that Bob rejected \textvisiblespace\ (immediately) &  \\
$b$ & & Bob left the room without closing the window &  \\
$c$ & ${}^{*}$ & window that Bob left the room without closing \textvisiblespace &  \\
$d$ & & papers that Bob rejected \textvisiblespace\ without reading \textvisiblespace$_p$\ (carefully)& \\
$e$ & & security breach that a report about \textvisiblespace$_p$\ in the NYT made \textvisiblespace\ public & \\
$f$ & & this is a candidate whom I would persuade every friend of \textvisiblespace\ to vote for \textvisiblespace\ & \\
\end{tabular}
}
\end{equation}

Consider first the case of object relativisation in ($a$). This example has a single gap for the unexpressed
direct object of \emph{rejected}. In categorial type logics, gaps have the status of \emph{hypotheses}, introduced
by a higher-order type. In Lambek's \cite{lam58} Syntactic Calculus, for example, the relative pronoun \emph{that}
in ($a$) would be typed as $(n\bs n)/(s/np)$. The complete relative clause then acts as a noun postmodifier $n\bs n$.
The relative clause body \emph{Bob rejected \textvisiblespace\ } is typed as $s/np$, which means it
needs a noun phrase hypothesis in order to compose a full sentence. Because the hypothesis occupies the
direct object position, it is impossible to physically realize that object, as the ungrammaticality of
${}^*$\emph{papers that Bob rejected the proposal} shows. The Lambek type requires the hypothetical $np$
to occur at the right periphery of the relative clause body --- a restriction that we will lift in
Section \S\ref{sec:syntax} to allow for phrase-internal hypotheses. An example would be ($a$) with an
extra temporal adverb (\emph{immediately}) at the end.

As the name suggests, a \emph{parasitic} gap is felicitous only in the presence of a primary gap.
The relative clause in ($d$) has two gaps: the primary one is for the object of
\emph{rejected} as in ($a$); the secondary, parasitic gap (marked by \textvisiblespace{}$_p$) is
the unexpressed object of \emph{reading}.
The parasitic gap occurs here in an adjunct: the verb phrase modifier \emph{without closing \textvisiblespace}.
Such an adjunct by itself, is an \emph{island} for extraction: the ungrammatical ($c$) shows that it
is impossible for the relative pronoun to establish communication with a $np$ hypothesis occuring within
the adjunct phrase. Compare ($c$) with the gapless ($b$) which has the complete adjunct
\emph{without closing (the window)$_{np}$}.
 
Examples ($e$) and ($f$) represent a different type of parasitic gapping
where both the primary and the parasitic gap regard co-arguments of the same verb.
In ($e$), the primary gap is the direct object of \emph{made public}, the secondary gap occurs
in the subject argument of this predicate. In ($f$), the primary gap is the object of the
infinitive complement of the verb \emph{persuade}, viz.~\emph{to vote for \textvisiblespace},
while the secondary gap occurs in the direct object of
\emph{persuade}\footnote{According to \cite{steedman87}, each of the gaps in this type of example would
be felicitous by itself.}.

We illustrated the adjunct and co-argument types of parasitic gapping in (\ref{data}) with relative clause examples.
Primary gaps can also be triggered in main or subordinate constituent question constructions,
as in (\ref{moredata}$a,b$), where \emph{which papers} will carry the higher-order type initiating
hypothetical reasoning. In the `passive infinitive' case (\ref{moredata}$c$), the higher-order
type is associated with the adjective \emph{hard}, which in this context could be typed as $ap/(\mathit{to\_inf}/np)$.
The adjective then selects for an incomplete $to$-infinitive missing a $np$ hypothesis, the direct object in
(\ref{moredata}$c$). As with the relative clause example (\ref{data}$a$), putting a physically realized
$np$ in the position occupied by the hypothesis leads to ungrammaticality.
Again, as in (\ref{data}), the primary gaps here open the possibility for
parasitic gaps dependent on them as in (\ref{moredata}$d,e,f$). These examples
also illustrate some of the various forms the adjunct phrase can take: temporal modification
(\emph{before, after}), contrastive (\emph{despite}), etc.

\begin{equation}\label{moredata}
\mbox{\begin{tabular}{r@{\quad}r@{}l@{\qquad}l}
$a$ & & which papers did Bob reject \textvisiblespace\ (immediately) &  \\
$b$ & & I know which papers Bob will reject \textvisiblespace\ (immediately) & \\
$c$ & & this paper is hard to understand \textvisiblespace\ / ${}^*$the proposal & \\
$d$ & & which papers did Bob accept \textvisiblespace\ \emph{despite} not liking \textvisiblespace$_p$\ (really) &  \\
$e$ & & I know which papers Bob will reject \textvisiblespace\ \emph{before} even reading \textvisiblespace$_p$\ (cursorily) & \\
$f$ & & this paper is easy to explain \textvisiblespace\ well \emph{after} studying \textvisiblespace$_p$\ (thoroughly) & \\
\end{tabular}
}
\end{equation}

To account for the duplication of semantic content in parasitic gap constructions, existing categorial analyses
rely on explicit forms of syntactic copying. The CCG analysis of \cite{steedman87} rests on (a directional version of) the
\textbf{S} combinator of Combinatory Logic; the type-logical account of \cite{morrill2015computational,morrill2016logic}
adapts the $\mathop{!}$ modality of Linear Logic to implement a restricted form of the structural rule of Contraction.
These syntactic devices are hard to control: the CCG version of the \textbf{S} rule is constrained by
rule features; %non-logical side conditions;
the attempts to properly constrain Contraction easily lead to undecidabilty as shown in \cite{KanovichKS19}.

Our aim in this paper is to explore \emph{lexical polymorphism} as an alternative to syntactic copying.
The technique of polymorphic typing is standardly used in categorial grammars for chameleon words such as
\emph{and, but}. Rather than giving these words a single type, they
are assigned a type \emph{schema}, with different realizations depending on whether they are 
conjoining sentences, verb phrases, transitive verbs, etc.
Treating the adjunct phrases of (\ref{data}$d$) and (\ref{moredata}$d,e,f$) as
forms of \emph{subordinating} conjunction, we propose to similarly handle the adjunct type of parasitic gaps
by means of a polymorphic type schema for the heads \emph{without, despite, after}, etc.
In the co-argument type of parasitic gapping (\ref{data}$e$,$f$), a conjunctive interpretation is absent.
In this case, a polymorphic type schema for the relative pronouns \emph{that} or \emph{who(m)} allows us to generalize from
the single gap instance (\ref{data}$a$) to the multi-gap case (\ref{data}$e$). To obtain the derived relative pronoun
type from the basic assignment, we can rely on the same mechanisms that relate the basic type for \emph{without} etc
to the derived type needed for the parasitic gap examples.

Our analysis builds on the categorical Frobenius algebraic compositional distributional semantics of \cite{sadrzadeh2013frobenius,sadrzadeh2014frobenius},
combined with a multimodal extension of Lambek calculus as the syntactic front end, as in \cite{moortgat2017lexical}.
Our analysis provides further evidence that Frobenius algebra is a powerful tool to model the
internal dynamics of lexical semantics. 

\section{Syntax}
\label{sec:syntax}

\subsection{The logic \textbf{NL}$_\fdia$}
The syntactic front end for our analysis is the type logic \textbf{NL}$_\fdia$ of  \cite{moortgat1996multimodal} which extends Lambek's pure logic of residuation \cite{lam61} with modalities for structural control. The formula language is given by the following grammar ($p$ atomic):
\begin{equation}\label{formulas}
A,B \mathbin{::=} p \mid A\otimes B \mid A/B \mid A\bs B \mid \fdia A \mid \gbox A
\end{equation}
In \textbf{NL}$_\fdia$, types are assigned to \emph{phrases}, not to strings as in the more familiar Syntactic Calculus of \cite{lam58}, or its pregroup version \cite{Lambek97}. The tensor product $\otimes$ then is
a non-associative, non-commutative operation for putting phrases together; it has adjoints $/$ and $\bs$ expressing right and left incompleteness with respect to phrasal composition,
as captured by the residuation inferences (\ref{binary}). In addition to the binary family $/,\otimes,\bs$, the extended language has unary control modalities $\fdia,\gbox$ which again form a residuated pair with the inferences in (\ref{unary}).
\begin{equation}\label{binary}
\arrw{A}{C/B}\textrm{\ \ iff\ \ }\arrw{A\otimes B}{C}\textrm{\ \ iff\ \ }\arrw{B}{A\bs C}
\end{equation}
\begin{equation}\label{unary}
\arrw{\fdia A}{B}\textrm{\ \ iff\ \ }\arrw{A}{\gbox B}
\end{equation}
The modalities serve a double purpose, either \emph{licensing} reordering or restructuring that would otherwise be forbidden, or \emph{blocking} structural operations that otherwise would be applicable. To license  rightward extraction, as found in English long-range dependencies, we use the postulates in (\ref{xright}). Postulate $\alpha_\diamond$ is a controlled form of
associativity: the $\fdia$ marking licenses a rotation of the tensor formula tree that leaves the order of the
components $A,B,\fdia C$ unaffected. Postulate $\sigma_\diamond$ implements a form of controlled commutativity: here the internal structure of the tensor formula tree is unaffected, but the components $B$ and $\fdia C$ are exchanged.
\begin{equation}\label{xright}
\begin{array}{rl}
\alpha_\diamond: &\arrw{(A\tensor B)\tensor\fdia C}{A\tensor(B\tensor\fdia C)}\\
\sigma_\diamond: &\arrw{(A\tensor B)\tensor\fdia C}{(A\tensor\fdia C)\tensor B}\\
\end{array}
\end{equation}
To block these structural operations from applying, we use a pair of modalities $\rdia,\rbox$.
Phrases that qualify as syntactic islands are marked off by $\rdia$. The modal island
demarcation makes sure that the input conditions for $\alpha_\diamond,\sigma_\diamond$ do not arise. The island markers $\rdia,\rbox$ have no associated structural rules; their logical behaviour is fully
characterized by (\ref{unary}).

\begin{figure}
    \[\infer[]{\arr{1_A}{A}{A}}{}
\qquad
\infer[]{\arr{g\circ f}{A}{C}}{\arr{f}{A}{B} & \arr{g}{B}{C}}\]

\[
\infer[]{\arr{f\tensor g}{A\tensor C}{B\tensor D}}{\arr{f}{A}{B} & \arr{g}{C}{D}}
\]

\[
%\infer[]{\arr{f\tensor g}{A\tensor C}{B\tensor D}}{\arr{f}{A}{B} & \arr{g}{C}{D}}
%\qquad
\infer[]{\arr{f/g}{A/D}{B/C}}{\arr{f}{A}{B} & \arr{g}{C}{D}}
\qquad
\infer[]{\arr{f\bs g}{B\bs C}{A\bs D}}{\arr{f}{A}{B} & \arr{g}{C}{D}}
\]

\[\infer[]{\arr{\Diamond f}{\fdia A}{\fdia B}}{\arr{f}{A}{B}}
\qquad
\infer[]{\arr{\Box f}{\gbox A}{\gbox B}}{\arr{f}{A}{B}}\]

\[\arr{\textit{ev}^{\bs}_{A,B}}{A \tensor A \bs B}{B}
\qquad
\arr{\textit{co-ev}^{\bs}_{A,B}}{B}{A \bs (A \tensor B)}\]
\[\arr{\textit{ev}^{\slash}_{A,B}}{B \s A \tensor A}{B} 
\qquad
\arr{\textit{co-ev}^{\slash}_{A,B}}{B}{(B \tensor A) \s A}\]
\[\arr{\textit{ev}^{\gbox}_{A}}{\fdia \gbox A}{A}
\qquad
\arr{\textit{co-ev}^{\gbox}_{A}}{A}{\gbox \fdia A}\]
\[\arr{\xrighta}{(A\tensor B)\tensor\fdia C}{A\tensor(B\tensor\fdia C)}
\qquad
\arr{\xrightc}{(A\tensor B)\tensor\fdia C}{(A\tensor\fdia C)\tensor B}\]
    \caption{Do\v{s}en style axiomatisation of \modalNL.}
    \label{fig:dosen_nldia}
\end{figure}
\textbf{NL}$_{\diamond}$ derivations will be represented using the axiomatisation of Figure \ref{fig:dosen_nldia}, due to Do\v{s}en \cite{dovsen1992brief}. This axiomatisation
takes (Co)Evaluation as primitive arrows, and recursively generalizes these by means of Monotonicity.
It is routine to show that the residuation inferences of (\ref{binary}) and (\ref{unary}) become
derivable rules given the axiomatisation of Figure \ref{fig:dosen_nldia}.
To streamline derivations, we will make use of the derived residuation steps.
Also, we will freely use (Co)Evaluation and the structural postulates (\ref{xright}) in their \emph{rule} form,
by composing them with Transitivity ($\circ$).

\begin{comment}
% to Appendix
\[
\infer[]{\arrw{A\otimes B}{C}}{\arrw{A}{C/B}}
\quad\leadsto\quad
\infer[\circ]{\arrw{A\otimes B}{C}}{
\infer[\textsf{Mon}_{\otimes}]{\arrw{A\otimes B}{C/B \otimes B}}{
\arrw{A}{C/B} & \arrw{B}{B}} & 
\infer[\textit{ev}^{/}]{\arrw{C/B \otimes B}{C}}{}}\]

\[
\infer[]{\arrw{A}{C/B}}{\arrw{A\otimes B}{C}}
\quad\leadsto\quad
\infer[\circ]{\arrw{A}{C/B}}{
\infer[\textsf{CoApp}_{/}]{\arrw{A}{(A\otimes B)/B}}{} &
\infer[\textsf{Mon}_{/}]{\arrw{(A\otimes B)/B}{C/B}}{
\arrw{A\otimes B}{C} & \arrw{B}{B}}}\]
\end{comment}

\subsection{Graphical calculus for \modalNL}

Wijnholds \cite{wijnholds2017coherent} gives a coherent diagrammatic language for the non-associative Lambek Calculus \textbf{NL}; the generalisation to \textbf{NL} with control modalities is
straightforward, see Figure \ref{fig:dosen_diagrams}. In short, each connective is assigned two \emph{links} that either compose or decompose a type built with that connective. Links (and diagrams) can be put together granted that their in- and outputs coincide. This system has a full recursive definition, and is shown to be sound and complete (i.e. coherent) with respect to the categorical formulation of the Lambek Calculus, given a suitable set of graphical equalities (not discussed in the current paper).
%For the pictorially inclined, the diagrammatic calculus provides a simple and intuitive
%way to present proofs of \modalNL.

As an illustration, we present the derivation of the simple relative clause example (\ref{data}$a$)
in symbolic and diagrammatic form. For this case of
non-subject\footnote{Subject relative clauses, e.g.~\emph{paper that \textvisiblespace\ irritates Bob}, do not involve any structural reasoning. 
The relative pronoun for subject relatives can be typed simply as $(n\bs n)/(np\bs s)$.} relativisation, the relative pronoun \emph{that} is typed as a functor that produces
a noun modifier $n\bs n$ in combination with a sentence that contains an unexpressed $np$ hypothesis
(\emph{Bob rejected \textvisiblespace{} immediately}). The subtype for the gap is the modally
decorated formula $\fdia\gbox np$. 
\begin{figure}
\begin{center}
    \begin{tabular}{c@{\hskip 1em}c@{\hskip 1em}c}
		\textbf{Identity} & \textbf{Composition} & \textbf{$\tensor$ Monotonicity} \\
		{%
\beginpgfgraphicnamed{diagrams_gijs/identity}
\tikzstyle{every picture}=[baseline=(current bounding box).east,scale=0.75,node distance=5mm]
\tikzstyle{every node}=[scale=0.75]
\InputIfFileExists{diagrams_gijs/identity.tikz}{}{\input{./tikz/diagrams_gijs/identity.tikz}}
\tikzstyle{every picture}=[]
\tikzstyle{every node}=[]
\endpgfgraphicnamed} & 
		{%
\beginpgfgraphicnamed{diagrams_gijs/composition}
\tikzstyle{every picture}=[baseline=(current bounding box).east,scale=0.75,node distance=5mm]
\tikzstyle{every node}=[scale=0.75]
\InputIfFileExists{diagrams_gijs/composition.tikz}{}{\input{./tikz/diagrams_gijs/composition.tikz}}
\tikzstyle{every picture}=[]
\tikzstyle{every node}=[]
\endpgfgraphicnamed} &
        {%
\beginpgfgraphicnamed{diagrams_gijs/mon_tensor}
\tikzstyle{every picture}=[baseline=(current bounding box).east,scale=0.75,node distance=5mm]
\tikzstyle{every node}=[scale=0.75]
\InputIfFileExists{diagrams_gijs/mon_tensor.tikz}{}{\input{./tikz/diagrams_gijs/mon_tensor.tikz}}
\tikzstyle{every picture}=[]
\tikzstyle{every node}=[]
\endpgfgraphicnamed} \\
        & & \\
        \textbf{$\bs$ Monotonicity} & \textbf{$\bs$ Evaluation} & \textbf{$\bs$ Co-evaluation} \\
        {%
\beginpgfgraphicnamed{diagrams_gijs/mon_backslash}
\tikzstyle{every picture}=[baseline=(current bounding box).east,scale=0.75,node distance=5mm]
\tikzstyle{every node}=[scale=0.75]
\InputIfFileExists{diagrams_gijs/mon_backslash.tikz}{}{\input{./tikz/diagrams_gijs/mon_backslash.tikz}}
\tikzstyle{every picture}=[]
\tikzstyle{every node}=[]
\endpgfgraphicnamed} &
        {%
\beginpgfgraphicnamed{diagrams_gijs/application_backslash}
\tikzstyle{every picture}=[baseline=(current bounding box).east,scale=0.75,node distance=5mm]
\tikzstyle{every node}=[scale=0.75]
\InputIfFileExists{diagrams_gijs/application_backslash.tikz}{}{\input{./tikz/diagrams_gijs/application_backslash.tikz}}
\tikzstyle{every picture}=[]
\tikzstyle{every node}=[]
\endpgfgraphicnamed} &
        {%
\beginpgfgraphicnamed{diagrams_gijs/coapplication_backslash}
\tikzstyle{every picture}=[baseline=(current bounding box).east,scale=0.75,node distance=5mm]
\tikzstyle{every node}=[scale=0.75]
\InputIfFileExists{diagrams_gijs/coapplication_backslash.tikz}{}{\input{./tikz/diagrams_gijs/coapplication_backslash.tikz}}
\tikzstyle{every picture}=[]
\tikzstyle{every node}=[]
\endpgfgraphicnamed} \\
        & & \\
        \textbf{$\fdia$ Monotonicity} & \textbf{$\gbox$ Evaluation} & \textbf{$\gbox$ Co-evaluation} \\
        {%
\beginpgfgraphicnamed{diagrams_gijs/mon_diamond}
\tikzstyle{every picture}=[baseline=(current bounding box).east,scale=0.75,node distance=5mm]
\tikzstyle{every node}=[scale=0.75]
\InputIfFileExists{diagrams_gijs/mon_diamond.tikz}{}{\input{./tikz/diagrams_gijs/mon_diamond.tikz}}
\tikzstyle{every picture}=[]
\tikzstyle{every node}=[]
\endpgfgraphicnamed}
        &
        {%
\beginpgfgraphicnamed{diagrams_gijs/application_diamond}
\tikzstyle{every picture}=[baseline=(current bounding box).east,scale=0.75,node distance=5mm]
\tikzstyle{every node}=[scale=0.75]
\InputIfFileExists{diagrams_gijs/application_diamond.tikz}{}{\input{./tikz/diagrams_gijs/application_diamond.tikz}}
\tikzstyle{every picture}=[]
\tikzstyle{every node}=[]
\endpgfgraphicnamed} &
        {%
\beginpgfgraphicnamed{diagrams_gijs/coapplication_diamond}
\tikzstyle{every picture}=[baseline=(current bounding box).east,scale=0.75,node distance=5mm]
\tikzstyle{every node}=[scale=0.75]
\InputIfFileExists{diagrams_gijs/coapplication_diamond.tikz}{}{\input{./tikz/diagrams_gijs/coapplication_diamond.tikz}}
\tikzstyle{every picture}=[]
\tikzstyle{every node}=[]
\endpgfgraphicnamed} \\
    \end{tabular}
    \vspace{2em}
    \ \\
    \begin{tabular}{c@{\hskip 2em}c}
        \textbf{Controlled associativity $\xrighta$} & \textbf{Controlled commutativity $\xrightc$} \\
        {%
\beginpgfgraphicnamed{diagrams_gijs/structural_associativity}
\tikzstyle{every picture}=[baseline=(current bounding box).east,scale=0.75,node distance=5mm]
\tikzstyle{every node}=[scale=0.75]
\InputIfFileExists{diagrams_gijs/structural_associativity.tikz}{}{\input{./tikz/diagrams_gijs/structural_associativity.tikz}}
\tikzstyle{every picture}=[]
\tikzstyle{every node}=[]
\endpgfgraphicnamed} &     {%
\beginpgfgraphicnamed{diagrams_gijs/structural_commutativity}
\tikzstyle{every picture}=[baseline=(current bounding box).east,scale=0.75,node distance=5mm]
\tikzstyle{every node}=[scale=0.75]
\InputIfFileExists{diagrams_gijs/structural_commutativity.tikz}{}{\input{./tikz/diagrams_gijs/structural_commutativity.tikz}}
\tikzstyle{every picture}=[]
\tikzstyle{every node}=[]
\endpgfgraphicnamed} \\ 
    \end{tabular}
\end{center}
    \caption{Do\v{s}en style axiomisation of \modalNL\ with diagrams. Monotonicity and (co)evaluation laws for $\s$ are fully symmetrical to the given diagrams for $\bs$.}
    \label{fig:dosen_diagrams}
\end{figure}
The $\fdia$ marking allows it to cross phrase boundaries on its way to the phrase-internal position
adjacent to the transitive verb \emph{rejected}. At that point, the licensing $\fdia$ has done its
work, and can be disposed of by means of the $ev^{\Box}$ axiom $\arrw{\fdia\gbox np}{np}$, which provides the $np$ object required by the transitive verb \emph{rejected}.
For legibility, we use words instead of their types for the lexical assumptions in the derivation below.
The steps labeled $\ell$ indicate the lexical look-up.

\begin{equation}\label{simplegap}
\resizebox{\textwidth}{!}{
%\deduce{\textcolor{red}{\lambda y_{4}.((\textsc{paper} \  y_{4}) \wedge ((\textsc{immediately} \  (\textsc{rejected} \  y_{4})) \  \textsc{Bob}))}}{\deduce{$\rule{0pt}{2em}$}{
   \infer[\textit{ev}^{\bs}]{\mbox{paper}\otimes_{}(\mbox{that}\otimes_{}(\mbox{Bob}\otimes_{}(\mbox{rejected}\otimes_{}\mbox{immediately}))) \longrightarrow n}{
      \infer[\ell]{n}{\mbox{paper}}
   & 
      \infer[\textit{ev}^{/}]{\mbox{that}\otimes_{}(\mbox{Bob}\otimes_{}(\mbox{rejected}\otimes_{}\mbox{immediately})) \longrightarrow n \bs_{}n}{
         \infer[\ell]{(n \bs_{}n) /_{}(s /_{}\diamondsuit_{}\Box_{}np)}{\mbox{that}}
      & 
         \hspace{-1cm}\infer[\textit{res}_{/}]{\mbox{Bob}\otimes_{}(\mbox{rejected}\otimes_{}\mbox{immediately}) \longrightarrow s /_{}\diamondsuit_{}\Box_{}np}{
            \infer[\alpha_{\diamond}]{(\mbox{Bob}\otimes_{}(\mbox{rejected}\otimes_{}\mbox{immediately}))\otimes_{}\diamondsuit^{}\Box_{}np \longrightarrow s}{
               \infer[\sigma_{\diamond}]{\mbox{Bob}\otimes_{}((\mbox{rejected}\otimes_{}\mbox{immediately})\otimes_{}\diamondsuit^{}\Box_{}np) \longrightarrow s}{
                  \infer[\textit{ev}^{\bs}]{\mbox{Bob}\otimes_{}((\mbox{rejected}\otimes_{}\diamondsuit^{}\Box_{}np)\otimes_{}\mbox{immediately}) \longrightarrow s}{
                     \infer[\ell]{np}{\mbox{Bob}}
                  & 
                     \infer[\textit{ev}^{\bs}]{(\mbox{rejected}\otimes_{}\diamondsuit^{}\Box_{}np)\otimes_{}\mbox{immediately} \longrightarrow np \bs_{}s}{
                        \infer[\textit{ev}^{/}]{\mbox{rejected}\otimes_{}\diamondsuit^{}\Box_{}np \longrightarrow np \bs_{}s}{
                           \infer[\ell]{(np \bs_{}s) /_{}np}{\mbox{rejected}}
                        & 
                           \infer[\textit{ev}^{\Box}]{\diamondsuit^{}\Box_{}np \longrightarrow np}{}
                        }
                     & 
                        \infer[\ell]{(np \bs_{}s) \bs_{}(np \bs_{}s)}{\mbox{immediately}}
                     }
                  }
               }
            }
         }
      }
   }%}}
}
\end{equation}

\begin{figure}[h!]
\begin{center}
	    {%
\beginpgfgraphicnamed{infoflow_parasitic_gap_simple1}
\tikzstyle{every picture}=[baseline=(current bounding box).east,scale=0.65,node distance=5mm]
\tikzstyle{every node}=[scale=0.65]
\InputIfFileExists{infoflow_parasitic_gap_simple1.tikz}{}{\input{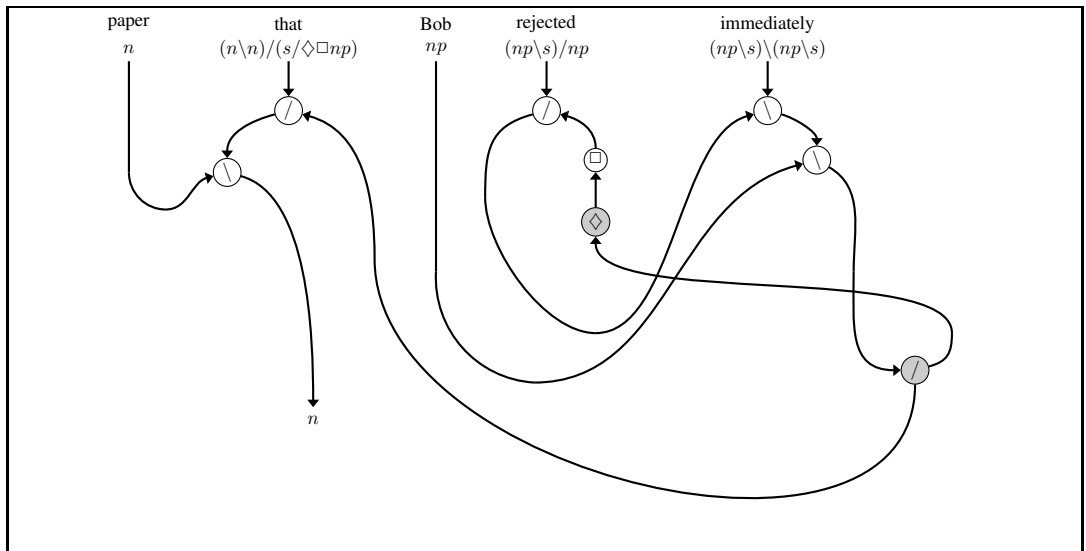}}
\tikzstyle{every picture}=[]
\tikzstyle{every node}=[]
\endpgfgraphicnamed}
	    \vspace{-3em}
\end{center}
\caption{Diagrammatic form of \emph{Paper that Bob rejected immediately}.}
\label{fig:immediately}
\end{figure}

In the diagrammatic form of Fig \ref{fig:immediately}, the $\diamondsuit \Box np$ gap hypothesis is indicated by the corresponding links. The leading $\fdia$ link licenses the crossing over to the object position of \emph{rejected} by means of the $\sigma_\diamond$ postulate of Fig \ref{fig:dosen_diagrams}.
In what follows, we use diagrams for \textbf{NL}$_{\diamond}$ derivations
because this format pictures the information flow in a simple and intuitive way.

\subsection{Typing Parasitic Gaps}

\paragraph*{Lexical polymorphism: generalized coordination} As our account of parasitic gaps in adjuncts treats the
adjuncts as a form of subordinate conjunction, we briefly review how lexical polymorphism is used in the analysis of generalized coordination.

Chameleon words such as \emph{and, but} cannot easily be typed monomorphically; given an initial type and interpretation, say $(s\bs s)/s$ for sentence coordination, we'd like to be able to obtain derived
types and interpretations for the coordination of (in)transitive verbs, as in (\ref{conj}$b,c$), or for non-constituent coordination cases such as (\ref{conj}$d$).
\newcommand{\dbn}{\fdia\gbox np}
\begin{equation}\label{conj}
\mbox{\begin{tabular}{r@{\quad}r@{}l@{\qquad}l}
$a$ & & (Alice sings)$_s$ and (Bob dances)$_s$ &  \\
$b$ & & Alice (sings and dances)$_{np\bs s}$ & \\
$c$ & & Bob (criticized and rejected)$_{(np\bs s)/np}$ the paper & \\
$d$ & & (Alice praised)$_{s/\dbn}$ but (Bob criticized)$_{s/\dbn}$ the paper & \\
\end{tabular}
}
\end{equation}
Deriving the ($b$--$d$) types from an initial $(s\bs s)/s$ assignment, however, goes beyond linearity. The attempt in (\ref{copy}) to derive verb phrase coordination from sentence coordination requires a copying step to strongly distribute the final $np$ abstraction over the two conjuncts.
\begin{equation}\label{copy}
\infer[]{\arrw{(s\bs s)/s}{((np\bs s)\bs(np\bs s))/(\fbox{$np$}\bs s)}}{
\infer[\textsf{Copy!}]{\arrw{\fbox{$np$}\otimes(np\bs s \otimes ((s\bs s)/s \otimes np\bs s)}{s}}{
\infer[]{\arrw{(\fbox{$np$}\otimes np\bs s)\otimes((s\bs s)/s \otimes(\fbox{$np$}\otimes np\bs s)}{s}}{\vdots}}}
\end{equation}
Partee and Rooth's \cite{partee-rooth83} work on generalized coordination offers a method
for replacing syntactic copying by lexical polymorphism.
Coordinating expressions \emph{and, but} get a polymorphic type assignment
$(X\bs X)/ X$ where $X$ is a conjoinable type. The set of conjoinable types CType forms
a subset of the general set of types Type. 
CType is defined inductively\footnote{Partee and Rooth formulate this in terms
of the semantic types obtained from the syntax-semantics homomorphism $h$,
with $h(s)=t$ (the type of truth values), $h(np)=e$ (individuals) and $h(A\bs B)=h(B/A)=h(A)\Ra h(B)$.}:

\medskip
\begin{itemize}[nosep]
  \item $s\in \textrm{CType}$;
  \item $A\bs B, B/A \in\textrm{CType}$ if $B\in\textrm{CType}$, $A\in\textrm{Type}$
\end{itemize}
\medskip
The type polymorphism comes with a generalized interpretation. We write $\sqcap^X$ (infix notation) for a coordinator of (semantic) type $X\Ra X\Ra X$. 

\medskip
\begin{itemize}[nosep]
  \item $P \sqcap^t Q \mathbin{:=} P\wedge Q$\quad coordination in type $t$ amounts to boolean conjunction
  \item $P \sqcap^{A\Ra B} Q \mathbin{:=} \lambda x^A.(P\ x)\sqcap^B (Q\ x)$ \quad distributing the $x^A$ parameter over the conjuncts
\end{itemize}
\medskip
The generalized interpretation scheme, then, associates a type transition such as (\ref{copy}) with the Curry-Howard program that would be associated with a derivation involving the copying step.
In Section \S\ref{sec:semantics}, we will obtain the same effect using the Frobenius algebras over our vector-based interpretations. 

\paragraph*{Parasitic gaps in adjuncts}
Consider the type lexicon for the data in
(\ref{data}$a$--$d$)\footnote{$iv$ abbreviates $np\bs s$; $gp$ stands for gerund clause, headed by the -ing form of the verb.}.
\begin{equation}\label{lexicon}
\begin{array}{r@{\quad::\quad}l}
\W{papers, window} & n \\
\W{that} & (n\bs n)/(s/\fdia\gbox np)\\
\W{Bob} & np \\
\W{rejected} & (np\bs s)/np \\
\W{reading, closing} & gp/np\\
\W{immediately, carefully} & iv\bs iv \\
\W{without} & \rbox(X\bs Y)/Z\qquad(\textrm{schematic})\\
\W{without}^{b,c} & \rbox(iv\bs iv)/gp\\
\W{without}^d & \rbox((iv/\fdia\gbox np)\bs (iv/ np))/(gp/\fdia\gbox np)\\
\end{array}
\end{equation}
The gap-less example (\ref{data}$b$) provides the motivation for the basic type assignment to \emph{without} as a functor combining
with a non-finite gerund clause $gp$ to produce a verb-phrase modifier $iv\bs iv$. To impose island constraints, we use a pair
of modalities $\rdia, \rbox$. In order to block the ungrammatical (\ref{data}$c$),
we follow \cite{morrillTLG} and lock the $iv\bs iv$ result type with $\rbox$; the matching $\rdia$ needed to unlock it has the effect of
demarcating the modifier phrase \emph{without closing the window} as an island, represented in the diagram below by means of a dotted line.
% \begin{figure}[h]
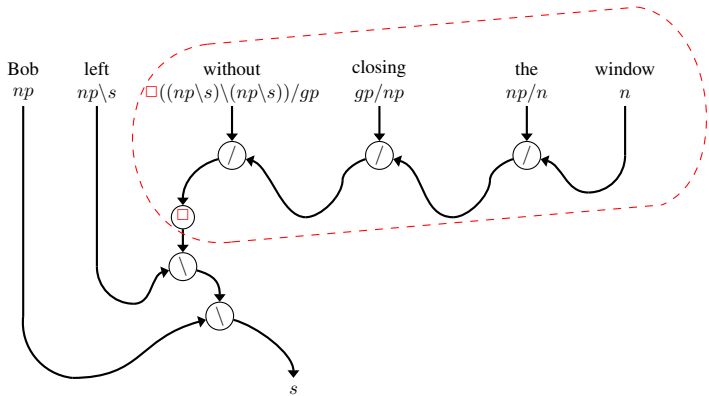
\begin{center}
	    {%
\beginpgfgraphicnamed{infoflow_parasitic_gap_simple2}
\tikzstyle{every picture}=[baseline=(current bounding box).east,scale=0.65,node distance=5mm]
\tikzstyle{every node}=[scale=0.65]
\InputIfFileExists{infoflow_parasitic_gap_simple2.tikz}{}{\input{./tikz/infoflow_parasitic_gap_simple2.tikz}}
\tikzstyle{every picture}=[]
\tikzstyle{every node}=[]
\endpgfgraphicnamed}
	   % \vspace{-1em}
\end{center}
% \caption{Diagrammatic form of \emph{Paper that Bob rejected immediately}.}
% \label{fig:infoflow_parasitic_gap}
% \end{figure}
% \input{Figures/parasitic_gap_simple2_derivation}
An attempt to derive the ungrammatical \emph{window that Bob left without closing \textvisiblespace\ } fails. The derivation proceeds like the one above, but with the gap hypothesis $\fdia\gbox np$
in the place of \emph{the window}. At that point the $\rdia$ island demarcation of \emph{without closing $\fdia\gbox np$} makes it impossible to bring out the hypothesis to the position where it can be withdrawn. This becomes apparent diagrammatically as the gap hypothesis cannot cross the dotted line:
\begin{figure}[h]
\begin{center}
	    {%
\beginpgfgraphicnamed{infoflow_parasitic_gap_simple3}
\tikzstyle{every picture}=[baseline=(current bounding box).east,scale=0.65,node distance=5mm]
\tikzstyle{every node}=[scale=0.65]
\InputIfFileExists{infoflow_parasitic_gap_simple3.tikz}{}{\input{./tikz/infoflow_parasitic_gap_simple3.tikz}}
\tikzstyle{every picture}=[]
\tikzstyle{every node}=[]
\endpgfgraphicnamed}
	    \vspace{-3em}
\end{center}
% \caption{Diagrammatic form of \emph{Paper that Bob rejected immediately}.}
\label{fig:infoflow_parasitic_gap}
\end{figure}
%
% \[\infer[]{\mbox{Bob}\otimes_{}(\mbox{left}\otimes_{}\rdia_{}(\mbox{without}\otimes_{}\mbox{closing}))\longrightarrow s/\fdia\gbox np}{
% \infer[]{(\mbox{Bob}\otimes_{}(\mbox{left}\otimes_{}\rdia_{}(\mbox{without}\otimes_{}\mbox{closing})))\otimes_{}\fdia\gbox np \longrightarrow s}{
% \infer[]{?}{
% \infer[]{\mbox{Bob}\otimes_{}(\mbox{left}\otimes_{}\rdia_{}(\mbox{without}\otimes_{}(\mbox{closing}\otimes_{}\fdia\gbox np))) \longrightarrow s}{\vdots}}}}\]
%

\noindent Let us turn then to the adjunct parasitic gapping of (\ref{data}$d$). To account for the double use of the gap we replace \emph{syntactic} copying via controlled Contraction by \emph{lexical} polymorphism,
treating \emph{without} as a polymorphic item on a par with coordinators \emph{and, but}. That means we assign to \emph{without}
the following type schema
\[\W{without}\mathbin{::}\rbox(X\bs Y)/Z\]
with basic instantiation $X=Y=iv$, $Z=gp$. From this basic instantiation, a derived instantiation with 
$X=Y=iv/\fdia\gbox np$ and $Z=gp/\fdia\gbox np$ is obtained for the parasitic gapping example (\ref{data}$d$) by uniformly dividing the subtypes $iv$ and $gp$ by $\fdia\gbox np$ using the forward slash.
\renewcommand{\dbn}{\fdia\gbox np}

In Section \S\ref{sec:semantics}, we will see how the vector-based interpretation of the derived type is obtained
in a systematic fashion from the interpretation of the basic type instantiation. For this,
it is helpful to factorize the construction of the derived type as the combination of an expansion step
and a distribution step. Ignoring the appropriate $\rbox$ decoration to mark off the adjunct as an island,
the expansion step here is an instance of the Geach transformation 
$\arrw{A/B}{(A/C)/(B/C)}$, with $A = iv\bs iv$, $B = gp$, $C = \dbn$.
\[\begin{tikzpicture}
\matrix (m) [matrix of math nodes, column sep=2em, row sep=3em]
{\textrm{(basic type)} & \rbox(iv\bs iv)/gp\\
 & (\rbox(iv\bs iv)/\dbn)/(gp/\dbn)\\
\textrm{(derived type)} & \rbox((iv/\dbn)\bs(iv/\dbn))/(gp/\dbn) \\
};
\path[->] (m-1-2) edge node [right] {expand} (m-2-2);
\path[->] (m-2-2) edge node [right] {distribute} (m-3-2);
\end{tikzpicture}
\]
Setting now $A=iv$, $B=iv$, $C=\dbn$, the distribution step is a directional instance of the $\mathbf{S}$ combinator $\arrw{(A\bs B)/C}{(A/C)\bs(B/C)}$.

To arrive at the version of the derived type for \emph{without} as we have it in our lexicon (\ref{lexicon}),
a final calibration is required. We replace the result type $iv/\fdia\gbox np$ by $iv/np$, dropping the modal marking required for controlled
associativity/commutativity. The final type $\rbox((iv/\dbn)\bs(iv/np))/(gp/\dbn)$ allows for the
derivation of the parasitic gapping example (\ref{data}$c$) displayed in Figure \ref{fig:infoflow_parasitic_gap}, 
but also for cases of Right Node Raising such as
\[\textrm{Bob (rejected without reading)$_{iv/np}$ all papers about linguistics}\]
where \emph{all papers about linguistics} is a plain $np$ rather than $\fdia\gbox np$.
\begin{figure}[h]
\begin{center}
        \vspace{-1em}
	    {%
\beginpgfgraphicnamed{infoflow_parasitic_gap}
\tikzstyle{every picture}=[baseline=(current bounding box).east,scale=0.55,node distance=5mm]
\tikzstyle{every node}=[scale=0.55]
\InputIfFileExists{infoflow_parasitic_gap.tikz}{}{\input{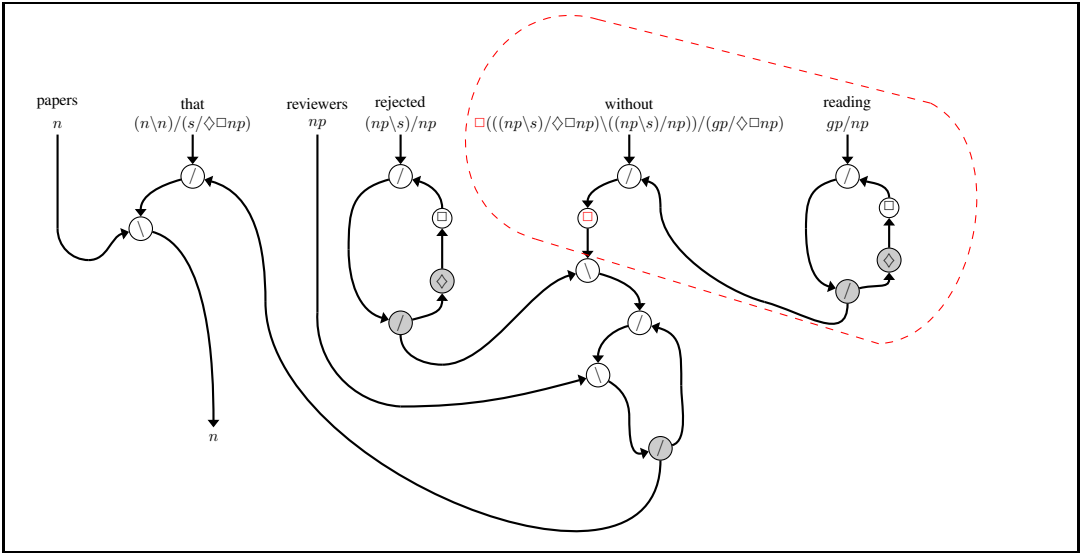}}
\tikzstyle{every picture}=[]
\tikzstyle{every node}=[]
\endpgfgraphicnamed}
	    \vspace{-4em}
\end{center}
\caption{Information flow for the double parasitic gap.}
\label{fig:infoflow_parasitic_gap}
\end{figure}

\subsubsection*{Parasitic gaps: co-arguments}
Let us turn to the co-argument type of parasitic gapping as exemplified by (\ref{data}$e,f$).
Consider first (\ref{data}$e$), repeated here for convenience, together with a gap-less
sentence that motivates the type-assignments given in (\ref{lex-co-arg-e}). 

\[\begin{array}{ll}
\mbox{security breach that a report about \textvisiblespace$_p$\ in the NYT made \textvisiblespace\ public} & =(1e)\\
\mbox{(a report in the NYT)$_{np}$ made (the security breach)$_{np}$ public$_{ap}$} & \\
\end{array}\]

\begin{equation}\label{lex-co-arg-e}
\begin{array}{r@{\quad::\quad}l}
\W{a, the} & np/n \\
\W{security breach, report, NYT} & n \\
\W{about, in} & (n\bs n)/np \\
\W{made} & ((np\bs s)/ap)/np \\
\W{public} & ap\\
\end{array}
\end{equation}

In (\ref{data}$e$) the relative clause body does not contain a coordination-like element that would be a suitable
candidate to lexically encapsulate the ostensible copying. But we can turn to the relative
pronoun itself, and use the mechanisms we relied on for parasitic gaps in adjuncts
to move from the relative pronoun's basic type assignment for single-gap dependencies to
a derived assignment for the double-gap dependency of (\ref{data}$e$).

\begin{equation}\label{lexthat}
\begin{array}{r@{\quad::\quad}l}
\W{that}^{a,c} & (n\bs n)/(s/\fdia\gbox np)\\
\W{that}^e & (n \bs_{}n) /_{}((np /_{}\diamondsuit_{}\Box_{}np) \otimes_{}((np \bs_{}s) /_{}\diamondsuit_{}\Box_{}np))
\end{array}
\end{equation}

Again, we see that these types are derivable from the initial type for \emph{that} by a combination of an expansion and a distribution step:
\[\begin{tikzpicture}
\matrix (m) [matrix of math nodes, row sep=3em]
{(n\bs n)/(s/\dbn)\\
(n\bs n)/((np\otimes np\bs s)/\dbn)\\
(n\bs n)/((np/\dbn)\otimes ((np\bs s)/\dbn)) \\
};

\path[->] (m-1-1) edge node [right] {expand} (m-2-1);
\path[->] (m-2-1) edge node [right] {distribute} (m-3-1);
\end{tikzpicture}
\]
The expansion step replaces $s$ in antitone position by $np\otimes np\bs s$, which is
justified by leftward Application $ev^{\bs}: \arrw{np\otimes np\bs s}{s}$ and Monotonicity. Here, with $A= np\otimes np\bs s$,
$B=s$, $C=\dbn$ and $D= n\bs n$, we have
\[\addtolength{\inferLineSkip}{2pt}
\infer[\W{Mon}^\downarrow]{\arrw{D/(B/C)}{D/(A/C)}}{
\infer[\W{Mon}^\uparrow]{\arrw{A/C}{B/C}}{
\infer[\W{Appl}]{\arrw{A}{B}}{}}}\]
Likewise, the distribution step relies on Mon$^\downarrow$ to replace $(A\otimes B)/C$ by $A/C\otimes B/C$ in
antitone position. Here, with $A=np$, $B=np\bs s$, $C=\dbn$, $D=n\bs n$, we have

\[\addtolength{\inferLineSkip}{2pt}
\infer[\W{Mon}^\downarrow]{\arrw{D/((A\otimes B)/C)}{D/(A/C\otimes B/C)}}{
\infer[\W{Res}]{\arrw{A/C\otimes B/C}{(A\otimes B)/C}}{
\infer[\W{Distr}]{\arrw{(A/C\otimes B/C)\otimes C}{A\otimes B}}{
\infer[]{\arrw{(A/C\otimes C)\otimes(B/C\otimes C)}{A\otimes B}}{\vdots}}}}\]
Figure \ref{coarg} has the derivation for example (\ref{data}$e$).
\begin{sidewaysfigure}
    \begin{center}
        {%
\beginpgfgraphicnamed{infoflow_parasitic_subject_en}
\tikzstyle{every picture}=[baseline=(current bounding box).east,scale=0.6,node distance=5mm]
\tikzstyle{every node}=[scale=0.6]
\InputIfFileExists{infoflow_parasitic_subject_en.tikz}{}{\input{./tikz/infoflow_parasitic_subject_en.tikz}}
\tikzstyle{every picture}=[]
\tikzstyle{every node}=[]
\endpgfgraphicnamed}     
    \end{center}
    \caption{Co-argument parasitic gapping (\ref{data}$e$).}
    \label{coarg}
\end{sidewaysfigure}

\begin{sidewaysfigure}
    \begin{center}
        {%
\beginpgfgraphicnamed{infoflow_parasitic_candidate_compact}
\tikzstyle{every picture}=[baseline=(current bounding box).east,scale=0.6,node distance=5mm]
\tikzstyle{every node}=[scale=0.6]
\InputIfFileExists{infoflow_parasitic_candidate_compact.tikz}{}{\input{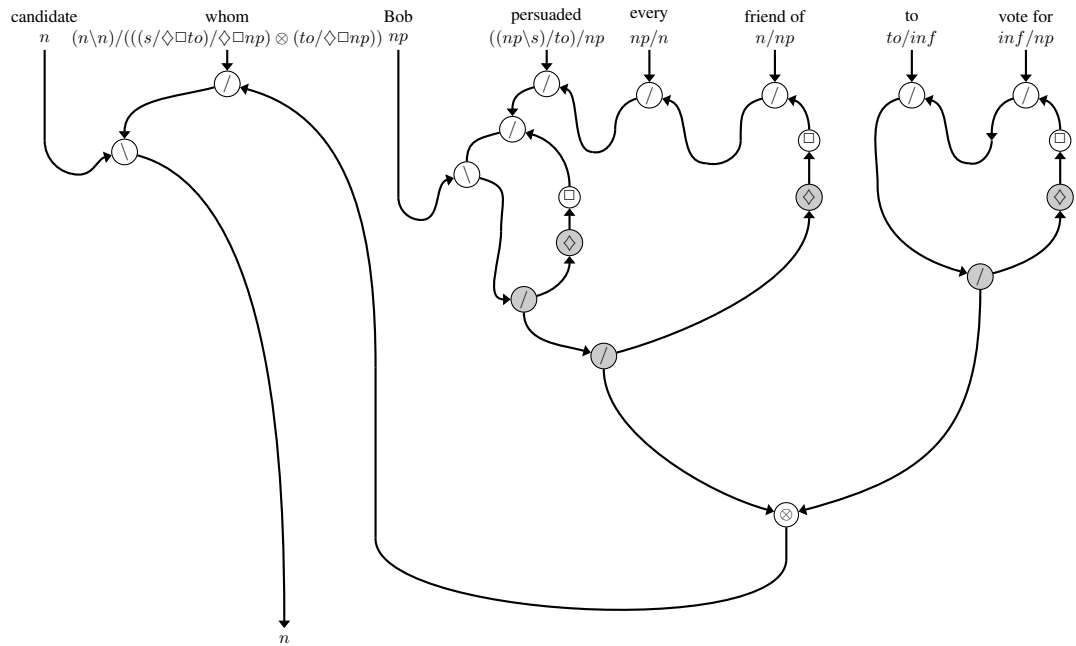}}
\tikzstyle{every picture}=[]
\tikzstyle{every node}=[]
\endpgfgraphicnamed}     
    \end{center}
    \caption{Co-argument parasitic gapping (\ref{data}$f$). For brevity we write \emph{to} for $\mathit{to\_inf}$.}
    \label{coarg}
\end{sidewaysfigure}

Turning to (\ref{data}$f$), repeated below with its underlying lexical type-assignments,
we find the primary and secondary gaps in the infinitival complement $\mathit{to\_inf}$
and direct object of the verb \emph{persuade}.

\[\begin{array}{ll}
\mbox{candidate whom Alice persuaded every friend of \textvisiblespace\ to vote for \textvisiblespace\ } & \sim(1f)\\
\mbox{Alice$_{np}$ persuaded (a friend)$_{np}$ to vote for Bob$_{np}$} & \\
\end{array}\]

\begin{equation}\label{lex-co-arg-f}
\begin{array}{r@{\quad::\quad}l}
\W{persuaded} & ((np\bs s)/\mathit{to\_inf})/np \\
\W{to vote} & \mathit{to\_inf}/pp\\
\W{for} & pp/np \\
\W{whom}^{f} & (n\bs n)/(((s/\fdia\gbox\mathit{to\_inf})/\dbn) \otimes (\mathit{to\_inf}/\dbn)) \\
\end{array}
\end{equation}

To obtain the required derived type for \emph{whom}, we follow the same expansion/distribution
routine as for (\ref{data}$f$). Expansion in this case replaces $s$ by the product
$(s/\mathit{to\_inf})\otimes \mathit{to\_inf}$; the gap type $\dbn$ is then distributed over
the two factors of that product. To obtain the desired $\W{whom}^{f}$, there is an extra modal
marking on the first occurrence of $\mathit{to\_inf}$, in order to license rebracketing with respect to the subject.
Recall that the base logic \textbf{NL} is non-associative by default.

\[\begin{tikzpicture}
\matrix (m) [matrix of math nodes, row sep=3em]
{(n\bs n)/(s/\dbn)\\
(n\bs n)/((s/\mathit{to\_inf})\otimes \mathit{to\_inf})/\dbn)\\
(n\bs n)/(((s/\mathit{to\_inf})/\dbn) \otimes (\mathit{to\_inf}/\dbn)) \\
};

\path[->] (m-1-1) edge node [right] {expand} (m-2-1);
\path[->] (m-2-1) edge node [right] {distribute} (m-3-1);
\end{tikzpicture}
\]

\section{Frobenius  Semantics}\label{sec:semantics}

The proposed vector-based semantics has two ingredients: first, the \emph{derivational} semantics
specifies a compositional mapping that interprets types and proofs of the \modalNL\ syntax as morphisms of a Compact Closed Category, concretely the category of \catvect\ and linear maps.
Second, the \emph{lexical} semantics specifies the word-internal interpretation of individual lexical items;
here, we make use of the Frobenius Algebras over \catvect\ to model the copying of semantic content associated with
the interpretation of relative pronouns such as \emph{that} and \emph{whom}, and modifier heads such as \emph{without}.

\subsection{Diagrams for Compact Closed Categories and Frobenius Algebras}

Recall that a Compact Closed Category is a symmetric  monoidal  category $({\cal C}, \otimes, I)$ with duals $A^*$ for every object $A$, and \emph{contraction} and \emph{expansion} maps for every object. In the case of vector spaces over fixed bases (our concrete semantics) we don't distinguish between objects and their duals, hence the contraction and expansion maps have signature $\epsilon : V \tensor V \pijl I$ and $\eta: I \pijl V \tensor V$, respectively.

For compact closed categories, there is a complete diagrammatic language available, that uses \emph{cups} and \emph{caps} to represent contraction and expansion, see \cite{Selinger2011}. These are drawn as connecting two objects either as a cup in the case of $\epsilon$ or as a cap in the case of $\eta$. The standard contraction and expansion maps of a CCC form the basis for interpreting derivations of \modalNL. 

\begin{figure}
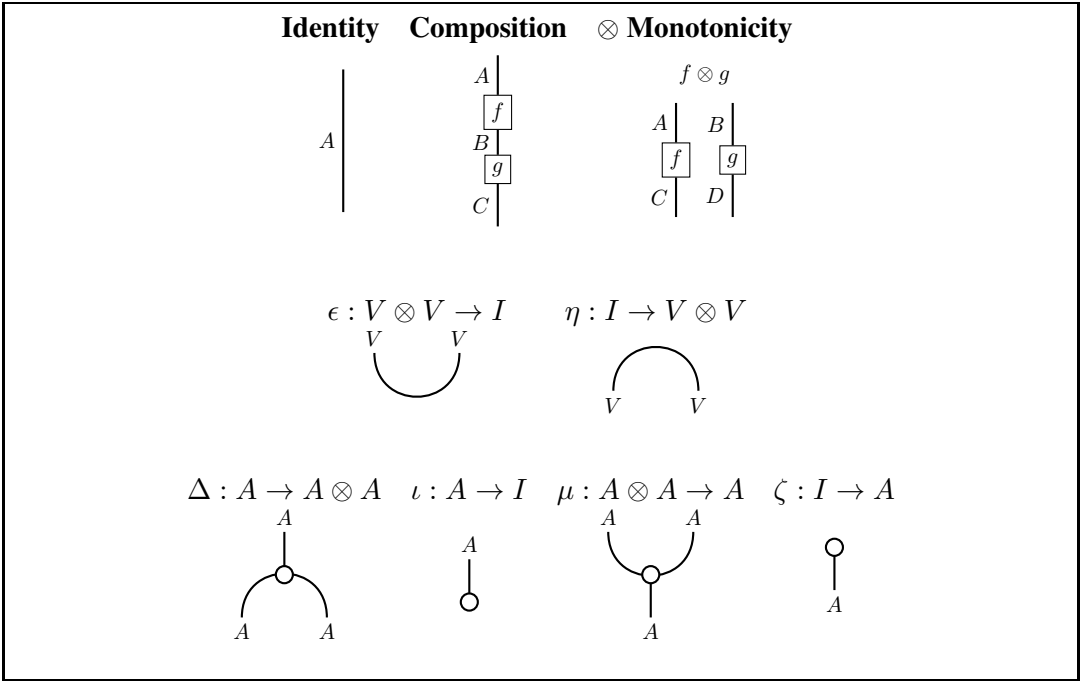

\begin{center}
    \begin{tabular}{c@{\hskip 1em}c@{\hskip 1em}c}
		\textbf{Identity} & \textbf{Composition} & \textbf{$\tensor$ Monotonicity} \\
		{%
\beginpgfgraphicnamed{diagrams_gijs/ccc_identity}
\tikzstyle{every picture}=[baseline=(current bounding box).east,scale=0.75,node distance=5mm]
\tikzstyle{every node}=[scale=0.75]
\InputIfFileExists{diagrams_gijs/ccc_identity.tikz}{}{\input{./tikz/diagrams_gijs/ccc_identity.tikz}}
\tikzstyle{every picture}=[]
\tikzstyle{every node}=[]
\endpgfgraphicnamed} & 
		{%
\beginpgfgraphicnamed{diagrams_gijs/ccc_composition}
\tikzstyle{every picture}=[baseline=(current bounding box).east,scale=0.75,node distance=5mm]
\tikzstyle{every node}=[scale=0.75]
\InputIfFileExists{diagrams_gijs/ccc_composition.tikz}{}{\input{./tikz/diagrams_gijs/ccc_composition.tikz}}
\tikzstyle{every picture}=[]
\tikzstyle{every node}=[]
\endpgfgraphicnamed} &
        {%
\beginpgfgraphicnamed{diagrams_gijs/ccc_tensor}
\tikzstyle{every picture}=[baseline=(current bounding box).east,scale=0.75,node distance=5mm]
\tikzstyle{every node}=[scale=0.75]
\InputIfFileExists{diagrams_gijs/ccc_tensor.tikz}{}{\input{./tikz/diagrams_gijs/ccc_tensor.tikz}}
\tikzstyle{every picture}=[]
\tikzstyle{every node}=[]
\endpgfgraphicnamed} \\
        \end{tabular}
        \vspace{2em}
        \ \\
        \begin{tabular}{c@{\hskip 2em}c}
        $\epsilon: V \tensor V \rightarrow I$ & $\eta : I \rightarrow V \tensor V$ \\
        {%
\beginpgfgraphicnamed{diagrams_gijs/cup}
\tikzstyle{every picture}=[baseline=(current bounding box).east,scale=0.75,node distance=5mm]
\tikzstyle{every node}=[scale=0.75]
\InputIfFileExists{diagrams_gijs/cup.tikz}{}{\input{./tikz/diagrams_gijs/cup.tikz}}
\tikzstyle{every picture}=[]
\tikzstyle{every node}=[]
\endpgfgraphicnamed} &
        {%
\beginpgfgraphicnamed{diagrams_gijs/cap}
\tikzstyle{every picture}=[baseline=(current bounding box).east,scale=0.75,node distance=5mm]
\tikzstyle{every node}=[scale=0.75]
\InputIfFileExists{diagrams_gijs/cap.tikz}{}{\input{./tikz/diagrams_gijs/cap.tikz}}
\tikzstyle{every picture}=[]
\tikzstyle{every node}=[]
\endpgfgraphicnamed} \\
        \end{tabular}
        \vspace{2em}
        \ \\
        \begin{tabular}{c@{\hskip 1em}c@{\hskip 1em}c@{\hskip 1em}c}
        $\Delta : A \pijl A \tensor A$ & $\iota : A \pijl I$ & $\mu : A \tensor A \pijl A$ & $\zeta : I \pijl A$ \\
        {%
\beginpgfgraphicnamed{diagrams_gijs/frob_delta}
\tikzstyle{every picture}=[baseline=(current bounding box).east,scale=0.75,node distance=5mm]
\tikzstyle{every node}=[scale=0.75]
\InputIfFileExists{diagrams_gijs/frob_delta.tikz}{}{\input{./tikz/diagrams_gijs/frob_delta.tikz}}
\tikzstyle{every picture}=[]
\tikzstyle{every node}=[]
\endpgfgraphicnamed}
        &
        {%
\beginpgfgraphicnamed{diagrams_gijs/frob_iota}
\tikzstyle{every picture}=[baseline=(current bounding box).east,scale=0.75,node distance=5mm]
\tikzstyle{every node}=[scale=0.75]
\InputIfFileExists{diagrams_gijs/frob_iota.tikz}{}{\input{./tikz/diagrams_gijs/frob_iota.tikz}}
\tikzstyle{every picture}=[]
\tikzstyle{every node}=[]
\endpgfgraphicnamed} &
        {%
\beginpgfgraphicnamed{diagrams_gijs/frob_mu}
\tikzstyle{every picture}=[baseline=(current bounding box).east,scale=0.75,node distance=5mm]
\tikzstyle{every node}=[scale=0.75]
\InputIfFileExists{diagrams_gijs/frob_mu.tikz}{}{\input{./tikz/diagrams_gijs/frob_mu.tikz}}
\tikzstyle{every picture}=[]
\tikzstyle{every node}=[]
\endpgfgraphicnamed} &
        {%
\beginpgfgraphicnamed{diagrams_gijs/frob_zeta}
\tikzstyle{every picture}=[baseline=(current bounding box).east,scale=0.75,node distance=5mm]
\tikzstyle{every node}=[scale=0.75]
\InputIfFileExists{diagrams_gijs/frob_zeta.tikz}{}{\input{./tikz/diagrams_gijs/frob_zeta.tikz}}
\tikzstyle{every picture}=[]
\tikzstyle{every node}=[]
\endpgfgraphicnamed} \\
        \end{tabular}
\end{center}
    \caption{Diagrams of a Compact Closed Category with Frobenius Algebras.}
    \label{fig:ccc_diagrams}
\end{figure}

Crucial to our polymorphic approach is the inclusion of Frobenius Algebras in the lexicon. A Frobenius algebra  in a  symmetric  monoidal  category $({\cal C}, \otimes, I)$ is a tuple $(X,  \Delta, \iota, \mu, \zeta)$ where,
for $X$ an object of ${\cal C}$, the first  triple below is  an internal comonoid and  the second one is  an internal monoid. 
\[
 (X, \Delta, \iota) \qquad (X, \mu, \zeta)
\]
This means that  we have a coassociative map $\Delta$ and and its counit $\iota$: 
\[
\Delta \colon X \to X \otimes X \qquad \iota \colon X \to I
\]
and an associative map $\mu$ and its unit  $\zeta$: 
\[
\mu \colon  X \otimes X \to X \qquad \zeta \colon I \to X
\]
as  morphisms of our category ${\cal C}$.
  \cut{
  \vspace{-0.2cm}
\begin{align*}
\Delta \colon X \to X \otimes X&\quad& \iota \colon X \to I &\qquad \qquad&\mu \colon  X \otimes X \to X  &\quad& \zeta \colon I \to X 
\end{align*}
Moreover  i.e.~the following are  associative and unital  morphisms:
\begin{align*}
\mu \colon  X \otimes X \to X  &\qquad& \zeta \colon I \to X
\end{align*}}
%\noindent The  $\Delta$ and $\mu$ morphisms satisfy a \emph{Frobenius condition}, which for reasons of space we do not review here. 
%\begin{align*}
%\mbox{\small $(\mu \otimes 1_X) \circ (1_X \otimes \Delta) \ = \  \Delta \circ \mu  \ = \  (1_X \otimes \mu) \circ (\Delta \otimes 1_X)$}
%\end{align*}
%MS
\noindent
The  $\Delta$ and $\mu$ morphisms satisfy the  \emph{Frobenius condition} given below\vspace{-1em}
\begin{align*}
\mbox{\small $(\mu \otimes 1_X) \circ (1_X \otimes \Delta) \ = \  \Delta \circ \mu  \ = \  (1_X \otimes \mu) \circ (\Delta \otimes 1_X)$}
\end{align*}
Informally, the comultiplication $\Delta$  decomposes the information contained in one object into two objects; the  multiplication $\mu$ combines the information of two objects into one. In diagrammatic terms, to visualise the Frobenius operations one adds a white triangle to the diagrammatic language for CCCs that represents the (un)merging of information through the four different Frobenius maps. The resulting graphical language is summarised in Figure \ref{fig:ccc_diagrams}.

\subsection{Derivational Semantics}

For the derivational semantics, we need to define a homomorphism $\hm{\cdot}$ that sends syntactic types and derivations to the corresponding components of the Compact Closed Category of \catvect\ and linear maps. This homomorphism has been worked out by Moortgat and Wijn\-holds \cite{moortgat2017lexical}. We present the key ingredients below and refer the reader to that paper for full details.

\noindent \textbf{Types} The target signature has atomic semantic spaces $N$ and $S$, an involutive $(\cdot)^{*}$ for dual spaces and a symmetric monoidal product $\otimes$.
We set
\begin{eqnarray*}
\hm{s} &=& S,\\
\hm{np}\, =\, \hm{n} &=&  N,\\
\hm{to\_inf}\, =\, \hm{ap}\, =\, \hm{gp} &=& N^{*}\otimes S,\\
 \hm{\fdia A} \,=\, \hm{\gbox A} &=& \hm{A},\\
 \hm{A/B} &=& \hm{A}\otimes\hm{B}^{*},\\
 \hm{A\bs B} &=& \hm{A}^{*}\otimes\hm{B}
\end{eqnarray*}
Notice that $to\_inf$, $ap$ and $gp$ are mapped to $N^{*}\otimes S$. Their understood subject is provided by the context: the main clause subject,
in the case of \emph{Bob fell asleep while watching TV}, the direct object in the case of \emph{make the report public} and \emph{persuade A to vote for B}.\\

\noindent \textbf{Derivations} The instances of the Evaluation axioms correspond to generalised contraction operations on vector spaces, the instances of the Co-Evaluation axioms dually are mapped to generalised expansion maps. The structural control postulates stipulate a syntactically limited associativity and commutativity; since the control modalities leave no trace on the semantic interpretation, the structural postulates $\alpha_\diamond$ and
$\sigma_\diamond$ are interpreted using the standard associativity and symmetry maps of \catvect.

\begin{figure}[h]
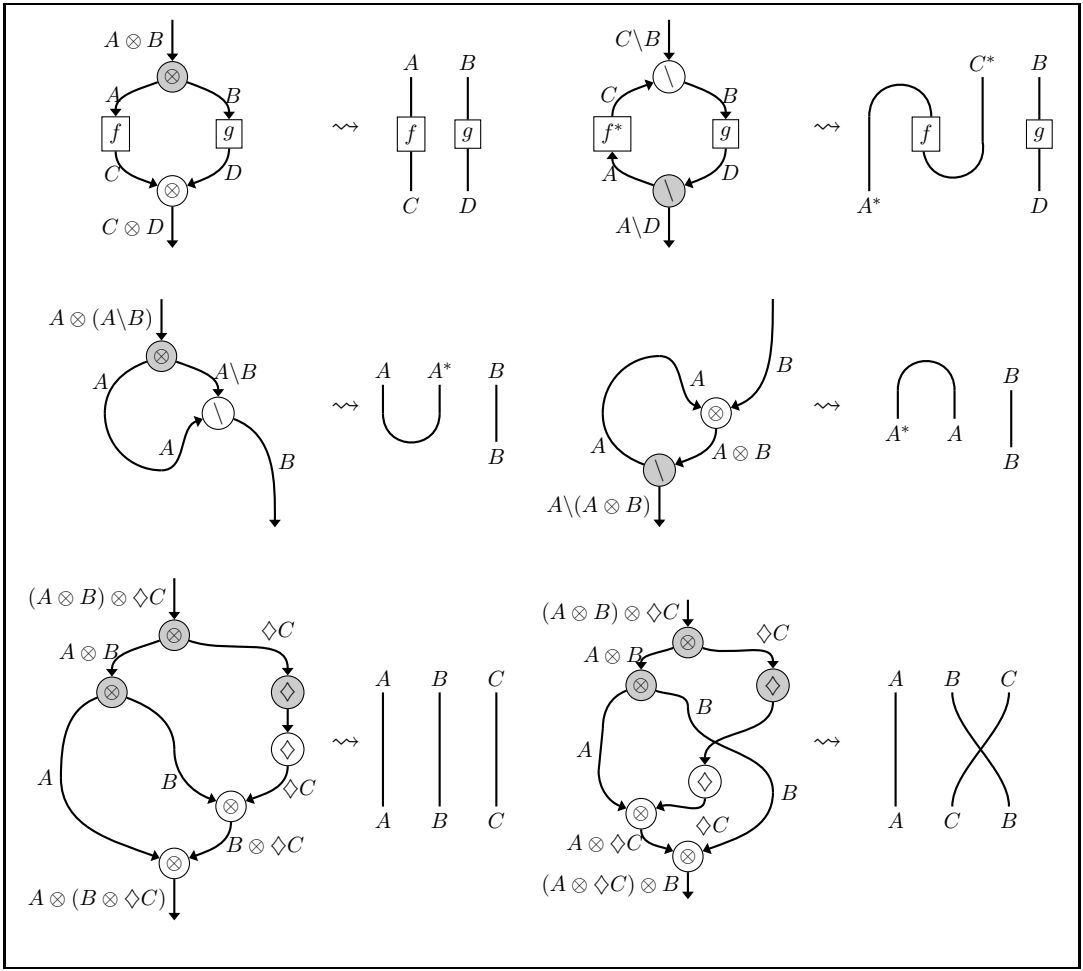

    \begin{center}
    \begin{tabular}{@{\hskip -0.75em}c@{\hskip -0.5em}c@{\hskip -0.5em}c@{\hskip -0.75em}c@{\hskip -0.5em}c@{\hskip -0.5em}c@{\hskip -0.5em}}
        {%
\beginpgfgraphicnamed{diagrams_gijs/mon_tensor}
\tikzstyle{every picture}=[baseline=(current bounding box).east,scale=0.75,node distance=5mm]
\tikzstyle{every node}=[scale=0.75]
\InputIfFileExists{diagrams_gijs/mon_tensor.tikz}{}{\input{./tikz/diagrams_gijs/mon_tensor.tikz}}
\tikzstyle{every picture}=[]
\tikzstyle{every node}=[]
\endpgfgraphicnamed} &
        $\rightsquigarrow$ &
        {%
\beginpgfgraphicnamed{diagrams_gijs/mon_tensor_ccc}
\tikzstyle{every picture}=[baseline=(current bounding box).east,scale=0.75,node distance=5mm]
\tikzstyle{every node}=[scale=0.75]
\InputIfFileExists{diagrams_gijs/mon_tensor_ccc.tikz}{}{\input{./tikz/diagrams_gijs/mon_tensor_ccc.tikz}}
\tikzstyle{every picture}=[]
\tikzstyle{every node}=[]
\endpgfgraphicnamed} &
        {%
\beginpgfgraphicnamed{diagrams_gijs/mon_backslash}
\tikzstyle{every picture}=[baseline=(current bounding box).east,scale=0.75,node distance=5mm]
\tikzstyle{every node}=[scale=0.75]
\InputIfFileExists{diagrams_gijs/mon_backslash.tikz}{}{\input{./tikz/diagrams_gijs/mon_backslash.tikz}}
\tikzstyle{every picture}=[]
\tikzstyle{every node}=[]
\endpgfgraphicnamed} &
        $\rightsquigarrow$ &
        {%
\beginpgfgraphicnamed{diagrams_gijs/mon_backslash_ccc}
\tikzstyle{every picture}=[baseline=(current bounding box).east,scale=0.75,node distance=5mm]
\tikzstyle{every node}=[scale=0.75]
\InputIfFileExists{diagrams_gijs/mon_backslash_ccc.tikz}{}{\input{./tikz/diagrams_gijs/mon_backslash_ccc.tikz}}
\tikzstyle{every picture}=[]
\tikzstyle{every node}=[]
\endpgfgraphicnamed} \\
        & \\
        {%
\beginpgfgraphicnamed{diagrams_gijs/application_backslash}
\tikzstyle{every picture}=[baseline=(current bounding box).east,scale=0.75,node distance=5mm]
\tikzstyle{every node}=[scale=0.75]
\InputIfFileExists{diagrams_gijs/application_backslash.tikz}{}{\input{./tikz/diagrams_gijs/application_backslash.tikz}}
\tikzstyle{every picture}=[]
\tikzstyle{every node}=[]
\endpgfgraphicnamed} & $\rightsquigarrow$ & {%
\beginpgfgraphicnamed{diagrams_gijs/application_ccc}
\tikzstyle{every picture}=[baseline=(current bounding box).east,scale=0.75,node distance=5mm]
\tikzstyle{every node}=[scale=0.75]
\InputIfFileExists{diagrams_gijs/application_ccc.tikz}{}{\input{./tikz/diagrams_gijs/application_ccc.tikz}}
\tikzstyle{every picture}=[]
\tikzstyle{every node}=[]
\endpgfgraphicnamed} &
        {%
\beginpgfgraphicnamed{diagrams_gijs/coapplication_backslash}
\tikzstyle{every picture}=[baseline=(current bounding box).east,scale=0.75,node distance=5mm]
\tikzstyle{every node}=[scale=0.75]
\InputIfFileExists{diagrams_gijs/coapplication_backslash.tikz}{}{\input{./tikz/diagrams_gijs/coapplication_backslash.tikz}}
\tikzstyle{every picture}=[]
\tikzstyle{every node}=[]
\endpgfgraphicnamed} & $\rightsquigarrow$ & {%
\beginpgfgraphicnamed{diagrams_gijs/coapplication_ccc}
\tikzstyle{every picture}=[baseline=(current bounding box).east,scale=0.75,node distance=5mm]
\tikzstyle{every node}=[scale=0.75]
\InputIfFileExists{diagrams_gijs/coapplication_ccc.tikz}{}{\input{./tikz/diagrams_gijs/coapplication_ccc.tikz}}
\tikzstyle{every picture}=[]
\tikzstyle{every node}=[]
\endpgfgraphicnamed} \\
        & \\
        {%
\beginpgfgraphicnamed{diagrams_gijs/structural_associativity}
\tikzstyle{every picture}=[baseline=(current bounding box).east,scale=0.75,node distance=5mm]
\tikzstyle{every node}=[scale=0.75]
\InputIfFileExists{diagrams_gijs/structural_associativity.tikz}{}{\input{./tikz/diagrams_gijs/structural_associativity.tikz}}
\tikzstyle{every picture}=[]
\tikzstyle{every node}=[]
\endpgfgraphicnamed} & $\rightsquigarrow$ & {%
\beginpgfgraphicnamed{diagrams_gijs/structural_associativity_ccc}
\tikzstyle{every picture}=[baseline=(current bounding box).east,scale=0.75,node distance=5mm]
\tikzstyle{every node}=[scale=0.75]
\InputIfFileExists{diagrams_gijs/structural_associativity_ccc.tikz}{}{\input{./tikz/diagrams_gijs/structural_associativity_ccc.tikz}}
\tikzstyle{every picture}=[]
\tikzstyle{every node}=[]
\endpgfgraphicnamed} &
        {%
\beginpgfgraphicnamed{diagrams_gijs/structural_commutativity}
\tikzstyle{every picture}=[baseline=(current bounding box).east,scale=0.75,node distance=5mm]
\tikzstyle{every node}=[scale=0.75]
\InputIfFileExists{diagrams_gijs/structural_commutativity.tikz}{}{\input{./tikz/diagrams_gijs/structural_commutativity.tikz}}
\tikzstyle{every picture}=[]
\tikzstyle{every node}=[]
\endpgfgraphicnamed} & $\rightsquigarrow$ & {%
\beginpgfgraphicnamed{diagrams_gijs/structural_commutativity_ccc}
\tikzstyle{every picture}=[baseline=(current bounding box).east,scale=0.75,node distance=5mm]
\tikzstyle{every node}=[scale=0.75]
\InputIfFileExists{diagrams_gijs/structural_commutativity_ccc.tikz}{}{\input{./tikz/diagrams_gijs/structural_commutativity_ccc.tikz}}
\tikzstyle{every picture}=[]
\tikzstyle{every node}=[]
\endpgfgraphicnamed} \\
    \end{tabular}
    \end{center}
    \caption{Interpreting derivations of \modalNL\  arrows in a compact closed category.}
    \label{fig:modalnl_to_ccc_diagrams}
\end{figure}

The derivational semantics is represented graphically in Figure \ref{fig:modalnl_to_ccc_diagrams}, where the diagrams of Figure \ref{fig:dosen_diagrams} are interpreted in the complete diagrammatic language of compact closed categories of Figure \ref{fig:ccc_diagrams}.

Under the given interpretation, the diagrammatic derivation of Figure \ref{fig:infoflow_parasitic_gap} for 
(\ref{data}$d$)
\[\scalebox{.9}{\begin{tabular}{lccccccc}
 & \W{papers} & \W{that} & \W{Bob}  & \W{rejected} & \quad \W{without} & \quad \W{reading} & \\
 & $n$ & $(n \bs n) / (s/ \fdia\gbox np)$ & $np$ & $(np\bs s)/np$ & \quad $(\rbox(X\bs Y))/Z$ & \quad $gp/np$ & $\longrightarrow n$
\end{tabular}}\]
is sent to the contractions in the interpreting CCC in Figure \ref{fig:linkings} (red: $\hm{\W{that}}$, blue: $\hm{\W{without}}$).
\input{tikz/linkings.tex}

% \begin{comment}

\subsection{Lexical Semantics}

\begin{figure}
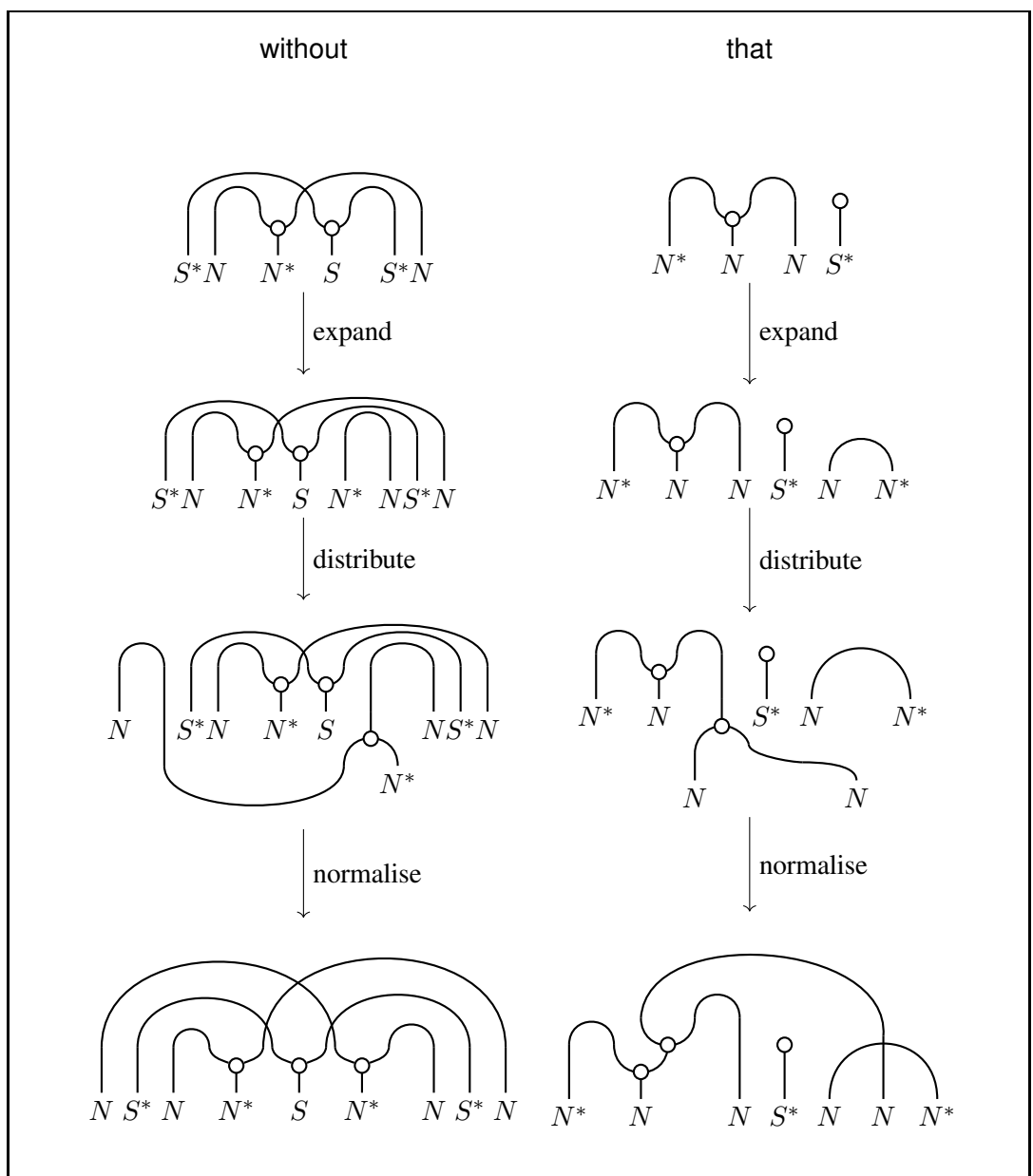

    \begin{center}
        \begin{tikzpicture}
            \matrix (m) [matrix of nodes, row sep=3em]
            {\textsf{without} & \textsf{that} \\
            {%
\beginpgfgraphicnamed{without_original}
\InputIfFileExists{without_original.tikz}{}{\input{./tikz/without_original.tikz}}
\endpgfgraphicnamed} & {%
\beginpgfgraphicnamed{that}
\InputIfFileExists{that.tikz}{}{\input{./tikz/that.tikz}}
\endpgfgraphicnamed}\\
            {%
\beginpgfgraphicnamed{without_expand}
\InputIfFileExists{without_expand.tikz}{}{\input{./tikz/without_expand.tikz}}
\endpgfgraphicnamed} & {%
\beginpgfgraphicnamed{that_expand}
\InputIfFileExists{that_expand.tikz}{}{\input{./tikz/that_expand.tikz}}
\endpgfgraphicnamed}\\
            {%
\beginpgfgraphicnamed{without_distribute2}
\InputIfFileExists{without_distribute2.tikz}{}{\input{./tikz/without_distribute2.tikz}}
\endpgfgraphicnamed} & {%
\beginpgfgraphicnamed{that_duplicate_1}
\InputIfFileExists{that_duplicate_1.tikz}{}{\input{./tikz/that_duplicate_1.tikz}}
\endpgfgraphicnamed}\\
            {%
\beginpgfgraphicnamed{without-long-coord}
\InputIfFileExists{without-long-coord.tikz}{}{\input{./tikz/without-long-coord.tikz}}
\endpgfgraphicnamed} & {%
\beginpgfgraphicnamed{that_duplicate_2}
\InputIfFileExists{that_duplicate_2.tikz}{}{\input{./tikz/that_duplicate_2.tikz}}
\endpgfgraphicnamed}\\
            };
            \path[->] (m-2-1) edge node [right] {expand} (m-3-1);
            \path[->] (m-3-1) edge node [right] {distribute} (m-4-1);
            \path[->] (m-4-1) edge node [right] {normalise} (m-5-1);
            \path[->] (m-2-2) edge node [right] {expand} (m-3-2);
            \path[->] (m-3-2) edge node [right] {distribute} (m-4-2);
            \path[->] (m-4-2) edge node [right] {normalise} (m-5-2);
        \end{tikzpicture}
    \end{center}
    \caption{Deriving the lexical semantics for \emph{without} and \emph{that}.}
    \label{fig:deriving_without_that}
\end{figure}

For the lexical interpretation of the relative pronouns \emph{that} and \emph{whom} and the conjunctive \emph{without}, we follow previous work \cite{sadrzadeh2013frobenius,sadrzadeh2014frobenius} and use Frobenius algebras that characterise  vector space bases \cite{CoeckeVic}. First, the basic form of the diagram for \emph{that} is as developed in \cite{sadrzadeh2013frobenius}. The basic diagram for \emph{without} uses a double instance of a Frobenius Algebra to coordinate the gerundive phrase with the intransitive verb phrase consumed to its left. 
Recall that the interpretation homomorphism sends $np\bs s$ and $gp$ to the same semantic space, $N^*\otimes S$. In Figure \ref{fig:deriving_without_that} we display graphically these basic types as well as how their \emph{derived} instantiations look. As our type for \emph{whom} is derived similarly to the type of \emph{that}, except that we distribute over the type $to\_inf$ rather than $np$, we get instead two extra wires rather than a single one.

For the case of parasitic gaps in adjunct positions we use the basic type for \emph{that} and the derived type for \emph{without}. For \emph{that}, its basic Frobenius instantiation has the concrete effect of projecting down the verb phrase into a vector which is consecutively multiplied element\-wise with the head noun of the main clause. The diagram for \emph{without} then makes sure to distribute the missing hypothesis of the relative clause over the two gaps in the clause body. Given the identification $\hm{iv}=\hm{gp}$, this is essentially the treatment of coordination of \cite{Kartsaklis16}.

For the co-argument case, we need make use of the derived type for \emph{that}; its function is now to both specify the need for a clause body missing a hypothetical noun phrase, as well as coordinating this noun phrase through two gaps. Hence, the derived instantiation figures an iterative use of the Frobenius $\mu$ to merge three elements together.
% The diagram for \emph{without} now coordinates the understood the three subjects and object, as well as the sentence types. Both diagrams are depicted below.\vspace{-1em}
% \begin{center}
% \begin{minipage}{4cm}
%   \tikzfig{that}
% \end{minipage}
% \qquad
% \begin{minipage}{6cm}
%   \tikzfig{without-long-coord}
% \end{minipage}
% \end{center}

\begin{figure}[h!]
\begin{center}
	    {%
\beginpgfgraphicnamed{infoflow_parasitic_gap_ccc_star}
\tikzstyle{every picture}=[baseline=(current bounding box).east,scale=0.55,node distance=5mm]
\tikzstyle{every node}=[scale=0.55]
\InputIfFileExists{infoflow_parasitic_gap_ccc_star.tikz}{}{\input{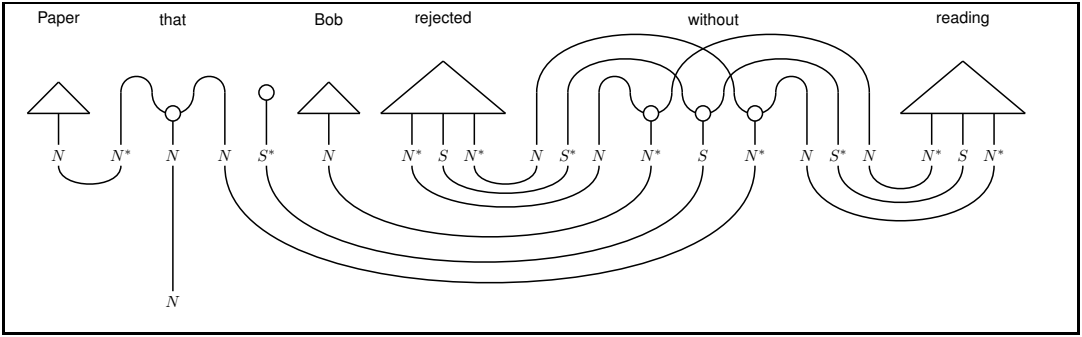}}
\tikzstyle{every picture}=[]
\tikzstyle{every node}=[]
\endpgfgraphicnamed}
	    \vspace{-1em}
\end{center}
\caption{Semantic information flow for the double parasitic gap (initial form).}
\label{fig:infoflow_parasitic_gap_ccc}
\end{figure}

With both the derivational semantics of Figure \ref{fig:linkings} and the lexical specifications of the constituents of Figure \ref{fig:deriving_without_that} we can put everything together to get the (unnormalised) diagram in Figure \ref{fig:infoflow_parasitic_gap_ccc}.

\begin{figure}[h]
\begin{center}
   {%
\beginpgfgraphicnamed{parasitic-norm-longer}
\InputIfFileExists{parasitic-norm-longer.tikz}{}{\input{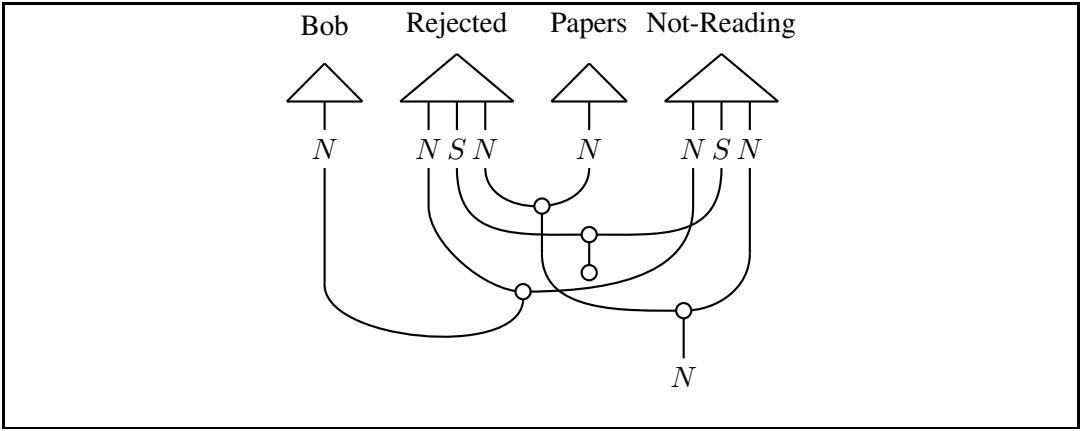}}
\endpgfgraphicnamed}
\end{center}
\caption{Semantic information flow for the double parasitic gap (normal form).}
\label{fig:infoflow_parasitic_gap_ccc_normal}
\end{figure}

This diagram can be normalised under the equations of the diagrammatic language, leading to the normal form of Figure \ref{fig:infoflow_parasitic_gap_ccc_normal}.

The above diagrams are morphisms of a symmetric compact closed category with Frobenius algebras  and can be written down in that language as done e.g.~in \cite{sadrzadeh2013frobenius,moortgat2017lexical}. Here, we provide the closed linear algebraic form of the normal form in Figure \ref{fig:infoflow_parasitic_gap_ccc_normal}. For $\overline{\overline{\text{Rejected}}}$ and  $\overline{\overline{\text{Not-Reading}}}$  the rank 3 tensors interpreting \emph{rejected} and \emph{(without) reading}, and $\iota$ the unit of the Frobenius coalgebra, this is
%\[
%\iota_S (\ov{\text{Reviewers}}^T \times \overline{\overline{\text{Rejected}}}) \odot (\ov{\text{Papers}} \odot (\iota_P \otimes id_N)(\overline{\text{Reading}}))
%\]
%
%The new normal form:
\[
 \ov{\text{Papers}} \odot (\iota_S \otimes id_N)  (\ov{\text{Bob}}^T \times
 (\overline{\overline{\text{Rejected}}} \odot 
\overline{\overline{\text{Not-Reading}}}))
\]

%\[
%(\ov{\text{Reviewers}}^T \times
%((id_N \otimes \iota_S \otimes id_N)\overline{\overline{\text{Rejected}}} %\odot 
%(id_N \otimes \iota_S \otimes id_N)\overline{\overline{\text{Reading}}})) %\odot \ov{\text{Papers}}
%\]

\noindent The closed linear algebraic form says that we take the elementwise multiplication of both cubes, and contract them with the subject \emph{Bob}; then, we collapse the resulting matrix into a vector and compute the elementwise multiplication of this vector with the vector interpreting the head noun \emph{Papers}.

For the co-argument case of parasitic gapping, we insert the derived Frobenius diagrams for \emph{that} and \emph{whom}, to obtain the initial diagrams of Figures \ref{fig:infoflow_parasitic_subject_en_full_ccc} (\ref{data}$e$) and \ref{fig:infoflow_parasitic_candidate_ccc_compact_v2}  (\ref{data}$f$), which normalise to the diagrams in Figures \ref{fig:infoflow_parasitic_subject_en_full_ccc_normal},\ref{fig:infoflow_parasitic_candidate_ccc_compact_v2_normal}. Note that the lexical specification of \emph{made} and \emph{persuade} is a wrapper around the lexical content of the verbs; since \emph{public} and the phrase \emph{to vote for} are interpreted as $N \otimes S$, their understood subject needs to be supplied, which happens through the use of Frobenius operations in the specification of their consuming verbs. This is the direct analogue of assigning a lambda term $\lambda x. \lambda P. \lambda y. \mathit{PERSUADE}\ x\ (P\ x)\ y$ to \emph{persuade}, where the Frobenius expansion corresponds to variable reuse.

\begin{sidewaysfigure}
\begin{center}
	    {%
\beginpgfgraphicnamed{infoflow_parasitic_subject_en_full_ccc_v2}
\tikzstyle{every picture}=[baseline=(current bounding box).east,scale=0.55,node distance=5mm]
\tikzstyle{every node}=[scale=0.55]
\InputIfFileExists{infoflow_parasitic_subject_en_full_ccc_v2.tikz}{}{\input{./tikz/infoflow_parasitic_subject_en_full_ccc_v2.tikz}}
\tikzstyle{every picture}=[]
\tikzstyle{every node}=[]
\endpgfgraphicnamed}
	    \vspace{-1em}
\end{center}
\caption{Semantic information flow for the co-argument parasitic gap (\ref{data}$e$, initial form).}
\label{fig:infoflow_parasitic_subject_en_full_ccc}
\end{sidewaysfigure}

\begin{sidewaysfigure}
\begin{center}
	    {%
\beginpgfgraphicnamed{infoflow_parasitic_candidate_ccc_compact_v2}
\tikzstyle{every picture}=[baseline=(current bounding box).east,scale=0.55,node distance=5mm]
\tikzstyle{every node}=[scale=0.55]
\InputIfFileExists{infoflow_parasitic_candidate_ccc_compact_v2.tikz}{}{\input{./tikz/infoflow_parasitic_candidate_ccc_compact_v2.tikz}}
\tikzstyle{every picture}=[]
\tikzstyle{every node}=[]
\endpgfgraphicnamed}
	    \vspace{-1em}
\end{center}
\caption{Semantic information flow for the co-argument parasitic gap (\ref{data}$f$, initial form).}
\label{fig:infoflow_parasitic_candidate_ccc_compact_v2}
\end{sidewaysfigure}

\begin{figure}[h!]
\begin{center}
	    {%
\beginpgfgraphicnamed{infoflow_parasitic_subject_en_full_ccc_5-2}
\InputIfFileExists{infoflow_parasitic_subject_en_full_ccc_5-2.tikz}{}{\input{./tikz/infoflow_parasitic_subject_en_full_ccc_5-2.tikz}}
\endpgfgraphicnamed}
\end{center}
\caption{Semantic information flow for the co-argument parasitic gap (\ref{data}$e$, normal form).}
\label{fig:infoflow_parasitic_subject_en_full_ccc_normal}
\end{figure}

\begin{figure}[h!]
\begin{center}
	    {%
\beginpgfgraphicnamed{infoflow_parasitic_candidate_ccc_compact_v2_normal}
\InputIfFileExists{infoflow_parasitic_candidate_ccc_compact_v2_normal.tikz}{}{\input{./tikz/infoflow_parasitic_candidate_ccc_compact_v2_normal.tikz}}
\endpgfgraphicnamed}
\end{center}
\caption{Semantic information flow for the co-argument parasitic gap (\ref{data}$f$, normal form).}
\label{fig:infoflow_parasitic_candidate_ccc_compact_v2_normal}
\end{figure}

\section{Discussion}

The concrete modelling presented above produces an interpretation of relative clauses that is analogous to the formal semantics account: seeing elementwise multiplication as an intersective operation (cf. set intersection), the interpretation of \emph{papers that Bob rejected without reading} identifies those papers that were both rejected and not reviewed, by Bob.

In the formal semantics account, the head noun and the relative clause body are both interpreted as functions from individuals to truth values, i.e.~characteristic functions of sets of individuals, which allows them to be combined by set intersection. In our vector-based modelling, however, the head noun and the relative clause body are initially sent to different semantic spaces, viz.~ $N$ for the head noun versus $N\otimes S$ for the relative clause body.
This means we need to appeal to the $\iota$ operation to effectuate the rank reduction from $N\otimes S$ to $N$ that reduces the
interpretation of the relative clause body to a vector that can then be conjoined with the meaning of the head noun. The rank reduction performed by the $\iota$ transformation is not a lossless transformation, and it is debatable whether it
correctly captures the semantic action we want to associate with the relative pronoun.

As a first step towards a more general model, we abstract away from the specific modelling of the relative pronoun by means of the $\iota$ map.

\begin{figure}[h]
\begin{center}
    {%
\beginpgfgraphicnamed{parasitic-norm-general}
\InputIfFileExists{parasitic-norm-general.tikz}{}{\input{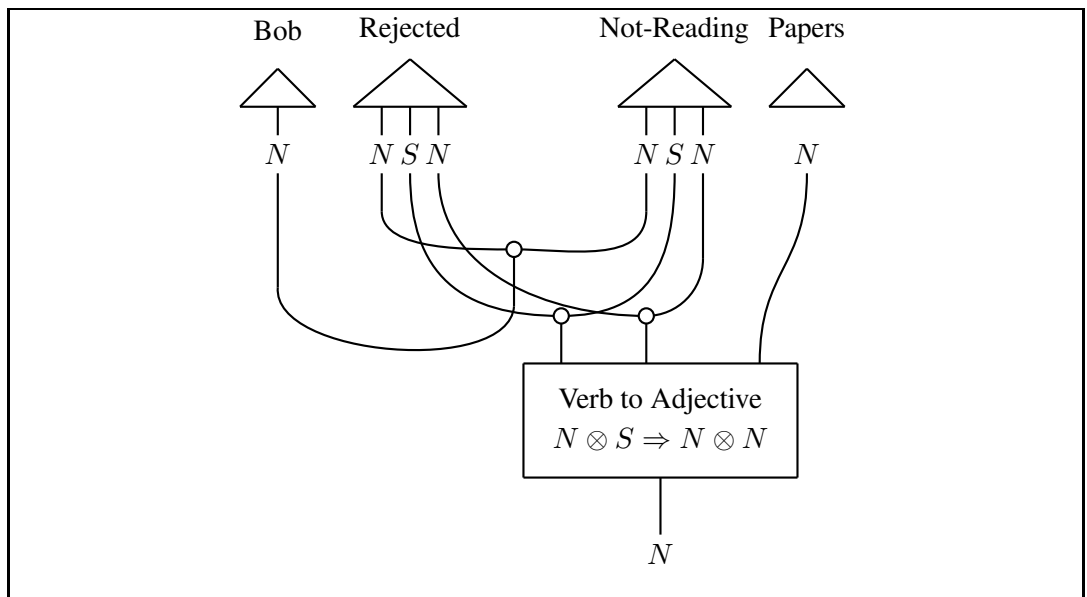}}
\endpgfgraphicnamed}
    \caption{General normal form for a sentence with a parasitic gap; the relative pronoun is now a general map that transforms a verb phrase ($N \tensor S$) into an adjective ($N\tensor N$).}
    \label{fig:norm_general}
\end{center}
\end{figure}

As shown in Figure \ref{fig:norm_general}, our type translation for the relative pronoun effectively interprets it as a map from a verb phrase ($N \tensor S$) meaning into an adjectival meaning modifying a (common) noun ($N \tensor N$).

With this generalization, we are not bound anymore to a specific implementation of the relative pronoun meaning, although the proposed account for now gives a workable solution for experimentation.

We suggest here, that a data-driven approach may lend itself for modelling the relative pronoun, as it essentially binds a verb phrase to its adjectival form. For example, a verb phrase can occur in adjectival form, e.g. ``papers that were rejected" vs ``rejected papers''. In such cases, we would expect to get the same meaning representation, which crucially relies on being able to project either an adjective onto a verb phrase or vice versa. Formulating this as a machine learning problem, is work in progress.
\section{Conclusion/Future Work}

We presented a typelogical ditributional account of parasitic gapping, one of the many linguistic phenomena in which some semantic elements are not  present in the sentence (or more generally discourse) and therefore their corresponding information needs to be provided from some other  syntactic element.
Rather than relying on some form of copying and/or movement on the syntax side to provide this information (as is the approach for ellipsis with anaphora in \cite{wijnholds2018classical,wijnholds2019typedriven}), we have solved the problem by using polymorphic typing for function words that play a key role in parasitic gapping (here, \emph{that}, \emph{whom} and \emph{without}).

The polymorphism carries over to the semantics, where  we have used  Frobenius algebras to interpret them. This enabled us to  handle the coordination of multiple gaps, and where the relative pronoun \emph{that} handles the coordination of the head noun with the body of the relative clause and the pronoun \emph{without} coordinates the second gap that exists in the body and which refers to the same head noun. The lexical specifications we use are analogous to a formal semantic modelling, but moreover allow for a more flexible way of representing meaning that may be obtained from data. That resolving gaps is useful in verb disambiguation and sentence similarity tasks has been recently shown \cite{wijnholds2019evaluating}. On this point, we discussed a more general normal form in which the behaviour of the relative pronoun is kept abstract. Investigating alternatives to the current modelling with the $\iota$ map, and looking into data-driven modelling of the relative pronoun, constitutes work in progress.

\bibliographystyle{classes/splncs03}
\bibliography{References/bibfile}
\nocite{Steedman87,partee-rooth83,Morrill19,KanovichKS19,Selinger2011}

% \newpage
% \appendix

% \input{Sections/BSyntaxDiagrams}
% \input{Sections/ASampleDerivations}

\end{document}